%% file: acl_latex.tex
\PassOptionsToPackage{table,dvipsnames}{xcolor}
\documentclass[11pt]{article}

\usepackage[final]{acl}
\usepackage{amsmath}
\usepackage{times}
\usepackage{latexsym}
\usepackage{seqsplit}
\usepackage{amssymb}
\usepackage{booktabs}
\usepackage{svg}
\usepackage{xcolor}
\usepackage{tikz}
\usepackage{caption}
\usetikzlibrary{arrows.meta,positioning}
\usepackage[T1]{fontenc}

\usepackage[utf8]{inputenc}

\usepackage{microtype}

\usepackage{inconsolata}

\usepackage{graphicx}
\usepackage{algorithm}
\usepackage{algorithmic}
\usepackage[most]{tcolorbox}
\newcommand{\mask}{\textcolor{gray!60}{\small [M]}}

\newcommand{\posdelta}[1]{%
  \,{\scriptsize\textcolor{ForestGreen}{$+#1$}}%
}
\newcommand{\negdelta}[1]{%
  \,{\scriptsize\textcolor{BrickRed}{$-#1$}}%
}

\title{Trace-Based On-Policy Distillation for Masked Diffusion Language Models}

\author{
  Haolin Ren\thanks{~~Equal contribution.}, Ziyang Huang$^*$, Chenhao Yuan, Jun Zhao, Kang Liu \\
  Institute of Automation, Chinese Academy of Sciences, Beijing, China \\
  University of Chinese Academy of Sciences, Beijing, China \\
  \texttt{renhaolin2026@ia.ac.cn}
}

\begin{document}
\maketitle

\input{0-abstract}
\input{1-introduction}
\input{2-background}
\input{3-method}
\input{4-experiment}
\input{5-relatedwork}
\input{6-conclusion}

\bibliography{custom}

\appendix

\input{7-appendix}

\end{document}

%% file: 0-abstract.tex
\begin{abstract}
    Diffusion large language models (dLLMs) are a promising alternative to autoregressive generation. However, reasoning-oriented post-training for dLLMs remains challenging. Supervised fine-tuning (SFT) for dLLMs requires dense but often off-policy masked states, while reinforcement learning (RL) relies on sparse rewards or value modeling. This paper proposes \textbf{trace-based on-policy distillation (TOPD)}, a teacher-supervised framework that transfers reasoning ability to a target dLLM without reward estimation. The key idea is to supervise a dLLM on its own denoising trajectory, focusing on the trace-aligned token decisions that form the final response. Specifically, TOPD samples on-policy diffusion trajectories from the target dLLM, obtains teacher token distributions from a teacher model on the corresponding partially denoised states, and updates the target dLLM with a token-level Reverse Kullback-Leibler (Reverse-KL) objective. This design preserves dense teacher supervision while aligning training with the model's own denoising states. On mathematical reasoning benchmarks, TOPD enables SDAR-4B-Chat to match the MATH500 accuracy of its RL-trained counterpart TraDo-4B-Instruct, with gains of +5.7 under static evaluation and +4.5 under dynamic evaluation. Compared with the RL-trained counterpart, TOPD achieves this with 4$\times$ fewer rollout rounds, corresponding to an estimated 96.0$\times$ to-accuracy model-compute speedup.
\end{abstract}

%% file: 1-introduction.tex
\section{Introduction}

\begin{figure}[t]
	\centering
	\includegraphics[width=\columnwidth]{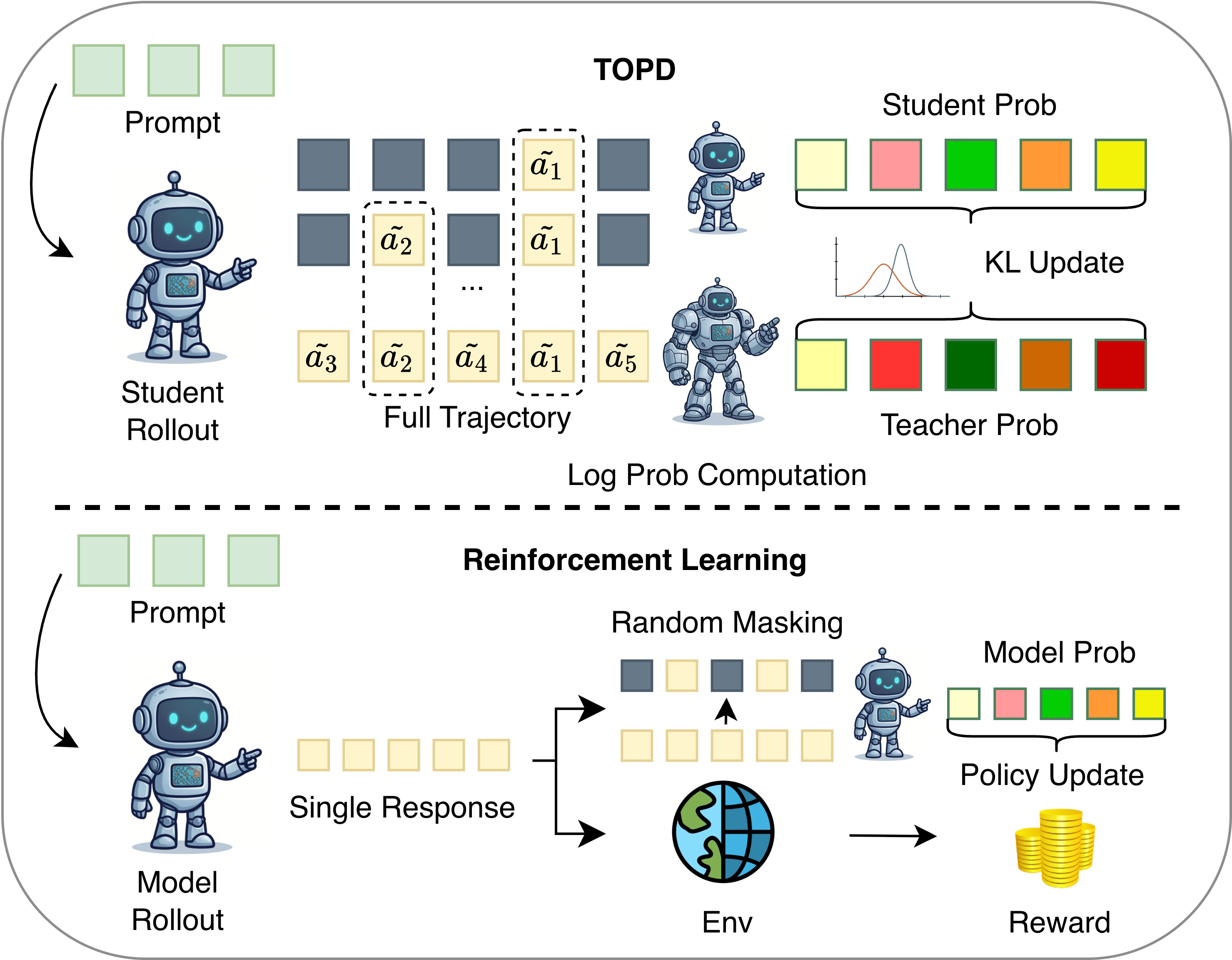}
		\caption{Conceptual comparison between trace-based on-policy distillation (TOPD) and random-mask reinforcement learning for dLLMs.}
	\label{fig:topd_comparison}
\end{figure}

\begin{figure}[t]
\centering
\resizebox{\columnwidth}{!}{%
\begin{tikzpicture}[
    font=\small,
    node distance=4mm,
    box/.style={draw, rounded corners=2pt, align=left, inner sep=4pt, minimum width=4.05cm, minimum height=0.82cm},
    msg/.style={draw, rounded corners=2pt, align=center, inner sep=4pt, text width=3.65cm, minimum height=0.82cm},
    good/.style={box, draw=ForestGreen!60!black, fill=ForestGreen!7},
    bad/.style={box, draw=BrickRed!70!black, fill=BrickRed!6},
    goodmsg/.style={msg, draw=ForestGreen!60!black, fill=ForestGreen!7},
    badmsg/.style={msg, draw=BrickRed!70!black, fill=BrickRed!6},
    full/.style={box, draw=black!45, fill=black!3, minimum width=8.7cm},
    note/.style={align=center, inner sep=2pt},
    arr/.style={-{Latex[length=1.6mm]}, thick}
]
\node[note] (qtitle) {\textbf{Question}};
\node[full, below=2mm of qtitle] (question) {A ticket costs \$80. It is discounted by 25\%,\\then taxed by 10\%. What is the final price?};
\node[note, below=4mm of question] (ftitle) {\textbf{Full answer}};
\node[full, below=2mm of ftitle] (target) {$80 \times (1-25\%) = 60$\\$60 \times (1+10\%) = 66$\\Final answer: \$66};

\node[note] (gtitle) at (-2.2,-4.05) {\textbf{On-policy state}};
\node[good, below=4mm of gtitle] (g1) {$80 \times (1-25\%) = \_$\\$\_ \times (1+10\%) = \_$\\Final answer: \$\_};
\node[goodmsg, below=of g1] (g2) {Preserves causal direction:\\subtotals are filled from\\preceding computations};
\draw[arr, ForestGreen!70!black] (target) -- (gtitle);
\draw[arr, ForestGreen!70!black] (g1) -- (g2);

\node[note] (btitle) at (2.2,-4.05) {\textbf{Random-mask state}};
\node[bad, below=4mm of btitle] (b1) {$\_ \times (1-25\%) = \_$\\$60 \times (1+\_\%) = 66$\\Final answer: \$\_};
\node[badmsg, below=of b1] (b2) {Breaks causal reasoning:\\60 is supervised from\\later evidence};
\draw[arr, BrickRed!75!black] (target) -- (btitle);
\draw[arr, BrickRed!75!black] (b1) -- (b2);
\end{tikzpicture}%
}
\caption{A synthetic question-answering example illustrating trajectory mismatch. Random masking can expose later results while hiding earlier reasoning variables, forcing backward reconstruction rather than forward computation.}
\label{fig:random_mask_mismatch}
\end{figure}
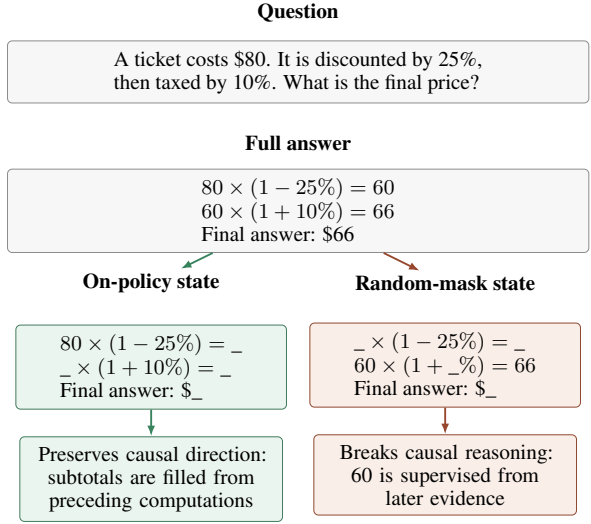

Post-training has driven large gains in autoregressive large language models (LLMs) on reasoning tasks. \citep{ouyang2022instructgpt,shao2024deepseekmath,guo2025deepseek,chu2025sft}. Diffusion large language models (dLLMs) offer a competitive non-autoregressive alternative based on iterative denoising \citep{li2022diffusionlm,sahoo2024simple,nie2025largelanguagediffusionmodels,cheng2025sdar,arriola2025block,gong2025scaling,ye2025dream7bdiffusionlarge}. However, reasoning-oriented post-training for dLLMs remains less well studied, particularly for distilling reasoning ability from stronger dLLMs into weaker student dLLMs. This paper studies how to efficiently adapt student dLLMs for reasoning under iterative denoising generation, using stronger dLLMs as supervision.

Current dLLM post-training pipelines mainly follow two routes: supervised fine-tuning (SFT) and reinforcement learning (RL) \citep{zhao2025d1,wang2025tracerl,zhu2025llada,he2025mdpo,ou2025espo,wang2025spg,zhao2025diffpo,zhong2026stabledrl}. 
SFT trains the student on fixed targets from external sources, such as human annotations and pre-generated samples, making the supervision inherently off-policy. 
In this way, the student is optimized on static targets or idealized teacher states rather than the states induced by its own generation policy, leading to exposure bias and weaker generalization \citep{agarwal2024policy,chu2025sft}. 
RL, on the other hand, often relies on sparse, delayed, and high-variance rewards, which makes long-horizon credit assignment difficult and optimization costly. 
These limitations motivate on-policy distillation (OPD) \citep{agarwal2024policy,lu2025onpolicydistillation}, where the student samples from its own policy while a teacher provides dense feedback on those student-generated outputs.

However, adapting OPD to dLLMs is not a direct transplant from autoregressive models. In autoregressive LLMs, each action is a committed next-token decision, so teacher feedback naturally attaches to the generated prefix. In dLLMs, the model iteratively updates a partially masked sequence, making trajectory alignment central. A key obstacle is the random-mask mismatch in many recent dLLM RL pipelines: training states are often built by randomly corrupting subsets of model-generated responses \citep{zhao2025d1,zhu2025llada,ou2025espo,wang2025spg,zhong2026stabledrl}. As illustrated in Figure~\ref{fig:random_mask_mismatch}, such states can expose downstream answers while hiding upstream reasoning variables, creating backward reconstruction contexts that differ from the student's forward denoising trajectory. Reveal order and inference schedules materially affect masked-diffusion behavior \citep{he2025mdpo,wang2025tracerl}. Teacher supervision on random or trajectory-agnostic states therefore weakens alignment with the decisions that form the final response.

To this end, this paper proposes \textbf{trace-based on-policy distillation (TOPD)}, a teacher-supervised post-training framework for dLLMs. TOPD samples the student's own inference-time denoising trajectories and converts them into dense distillation targets by identifying the token predictions that directly determine the final response. This trace-based formulation avoids supervising arbitrary random-mask configurations, which may expose later answer tokens while hiding earlier reasoning tokens and thus create conditional dependencies inconsistent with inference-time denoising. TOPD then uses a frozen teacher to provide token distributions on the student's on-policy partially denoised states and optimizes the student with step-wise token-distribution supervision. In this process, a token-level Reverse Kullback-Leibler (Reverse-KL) is exploited. As a result, TOPD preserves trajectory-aligned state coverage while replacing sparse-reward credit assignment and value-model overhead with dense teacher feedback. Figure~\ref{fig:topd_comparison} provides a conceptual comparison between TOPD and random-mask RL.

Experiments on mathematical reasoning benchmarks show that TOPD can effectively transfer reasoning ability from a dLLM teacher to a smaller student with substantially lower training cost. Specifically, using TraDo-8B-Instruct \citep{wang2025tracerl} as the teacher, TOPD brings the base SDAR-4B-Chat \citep{cheng2025sdar} student to the same MATH500 \citep{hendrycks2021measuring} accuracy as TraDo-4B-Instruct, an RL-trained version of SDAR-4B-Chat, with gains of +5.7 under static evaluation and +4.5 under dynamic evaluation. Moreover, compared with TraceRL \citep{wang2025tracerl}, TOPD achieves this with 4$\times$ fewer rollout rounds, yielding an estimated 96.0$\times$ to-accuracy model-compute speedup under our parameter-scaled accounting.

Our contributions are summarized as follows:
\begin{itemize}
    \item We formulate teacher-supervised dLLM post-training as a trace-based on-policy distillation problem, showing that random-mask supervision can induce conditional dependencies misaligned with the student's inference-time denoising trajectory.
    \item This paper proposes TOPD, an efficient post-training paradigm for dLLMs, which samples on-policy diffusion trajectories from the student and identifies the trace-aligned denoising decisions that directly determine the final response, replacing reward-based updates with step-wise teacher token-distribution supervision.
\end{itemize}

%% file: 2-background.tex
\section{Background}

\subsection{Masked Diffusion Language Models}

Masked diffusion language models (MDLMs) generate text by iteratively denoising masked token sequences rather than by committing tokens from left to right \citep{sahoo2024simple,nie2025largelanguagediffusionmodels}. Given a clean sequence $x\in\mathcal{V}^n$, training corrupts tokens into \texttt{[MASK]} at noise level $t$ and learns a denoiser $\pi_{\theta}(\cdot\mid z)$ for the masked positions, usually with a schedule-weighted masked reconstruction objective \citep{sahoo2024simple,shi2024simplified}. Full-attention dLLMs allow response tokens to attend bidirectionally to the whole corrupted response, while block-attention dLLMs partition the response into blocks and use a block-causal mask, e.g., a token in block $k$ attends only to blocks $\le k$ \citep{arriola2025block,cheng2025sdar}. At inference time, however, the model follows its own denoising trajectory $\tau=(s_0,a_0,\ldots,s_T)$ from a fully masked response, and the resulting reveal order can affect generation behavior \citep{he2025mdpo,wang2025tracerl}. Since intermediate decisions may later be revised or removed, we use \emph{trace} to denote the subset of decisions in $\tau$ that survives into $s_T$.

\subsection{Post-Training Signals for dLLMs}

Post-training for dLLMs typically relies on SFT or RL \citep{zhao2025d1,wang2025tracerl,zhu2025llada,he2025mdpo,ou2025espo,wang2025spg,zhao2025diffpo,zhong2026stabledrl}.  In SFT, a target response is partially corrupted into $\tilde{y}$ and the model is trained to reconstruct the masked positions:
\begin{equation}
    \mathcal{L}_{\mathrm{SFT}}(\theta)
    =
    \mathbb{E}_{q,y,\tilde{y}}
    \left[
    \sum_{i\in\mathcal{M}(\tilde{y})}
    -\log \pi_{\theta}(y_i\mid q,\tilde{y})
    \right],
    \label{eq:sft_masked}
\end{equation}
where $\mathcal{M}(\tilde{y})$ is the masked-position set and the expectation over $\tilde{y}$ denotes the random masking process applied to $y$. RL instead samples from the current student and maximizes
\begin{equation}
    J_{\mathrm{RL}}(\theta)
    =
    \mathbb{E}_{q,\,\tau\sim P_{\theta}(\cdot\mid q)}
    \bigl[R(q,s_T)\bigr].
\end{equation}
Here $P_{\theta}(\tau\mid q)$ denotes the trajectory distribution induced by the student's iterative denoising policy, and $R(q,s_T)$ is the scalar reward assigned to the final response. 
SFT provides dense supervision but trains on fixed target responses rather than states induced by the current student. RL uses student samples, but its sparse and delayed rewards must be assigned across many non-autoregressive denoising decisions, making credit assignment costly and high variance. Teacher supervision offers a middle ground: dense token distributions on student-visited states without delayed reward estimation.

\subsection{Challenges in Adapting OPD to dLLMs}

OPD trains a student on states sampled from its current policy while a teacher supplies dense target distributions on those same states \citep{agarwal2024policy,lu2025onpolicydistillation}. Adapting this idea to dLLMs is non-trivial because a diffusion step operates on a partially masked sequence and may propose multiple token decisions in parallel. Unlike autoregressive OPD, where each generated prefix defines a single committed next-token context, dLLM distillation must decide which partially denoised states and token positions should receive teacher feedback along a multi-step reveal process. Random-mask supervision provides dense masked-token targets, but these artificial states may not match the states or reveal order encountered by the current student during generation \citep{he2025mdpo,wang2025tracerl}.

The core challenge is therefore to make OPD both on-policy and trajectory-aware. The student should be supervised on states sampled from its own denoising process, and the teacher signal should be organized according to the trace construction induced by that process rather than by independently sampled masks. TOPD addresses this by sampling on-policy diffusion trajectories and distilling teacher distributions on trace-aligned denoising decisions, which we formalize next.

%% file: 3-method.tex
\section{Method}
\label{sec:methodology}

\begin{figure*}[t]
    \centering
    \includegraphics[width=\textwidth]{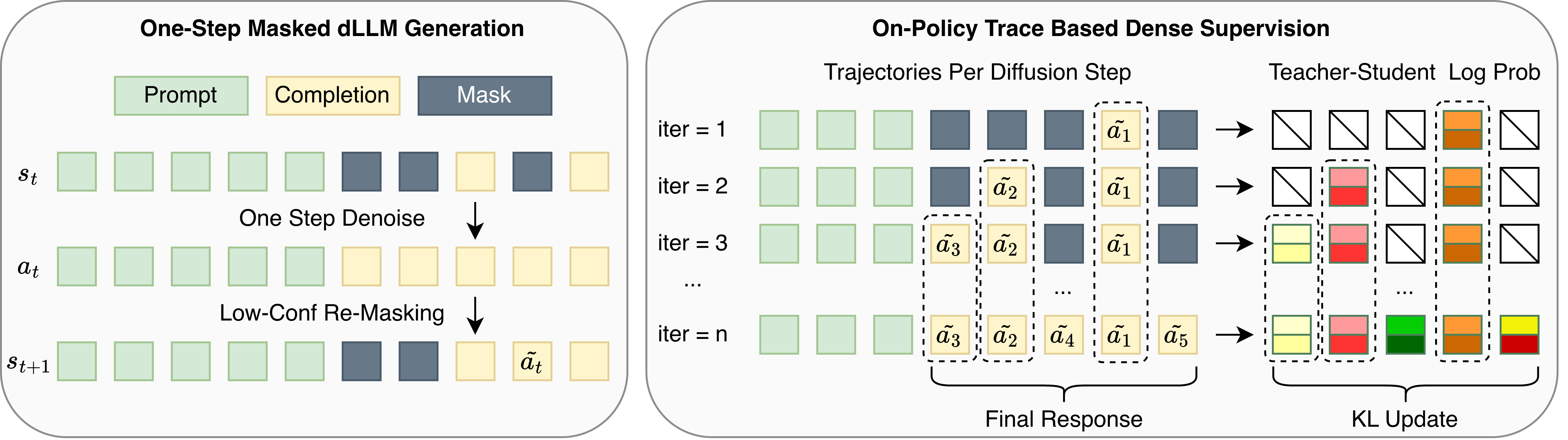}
    \caption{Overview of TOPD. The student samples an on-policy denoising trajectory, the teacher provides step-wise token-distribution supervision, and updates are applied only to trace-aligned decisions $\tilde{a}_t$ retained in the final output.}
    \label{fig:topd_full_pipeline}
\end{figure*}

We introduce \textbf{TOPD}, a trace-based on-policy distillation framework for post-training masked diffusion language models. Given a prompt, TOPD samples a student denoising trajectory, keeps the token decisions that survive into the final response, and distills a frozen teacher distribution on those student-visited states.

Figure~\ref{fig:topd_full_pipeline} summarizes the TOPD pipeline. Section~\ref{subsec:trajectory} defines the on-policy diffusion trajectory that supplies training states. Section~\ref{subsec:trace_alignment} describes how TOPD extracts trace-aligned decisions from that trajectory. Section~\ref{subsec:dense_supervision} constructs the teacher--student distribution pairs on the retained state-position pairs. Section~\ref{subsec:objective} gives the Reverse-KL objective and its sampled-token estimator.

\subsection{On-Policy Diffusion Trajectories}
\label{subsec:trajectory}

The first challenge is to supervise states that match the student's diffusion-time behavior. TOPD therefore samples trajectories from the current student rather than constructing states by randomly masking completed answers. Let $\pi_{\theta}$ denote the student dLLM and let $q$ be a prompt. Generation starts from a fully masked response state $s_0$ and proceeds for $T$ diffusion steps; at step $t$, the student predicts tokens for selected masked positions and updates the response to $s_{t+1}$. We record the rollout as
\begin{equation}
    \tau =
    (s_0, a_0, s_1, a_1, \ldots, s_{T-1}, a_{T-1}, s_T),
\end{equation}
where $s_T$ is the final response and $a_t=\{(j,x_j)\}$ is the set of token decisions proposed and accepted at step $t$.

Sampling $\tau$ from the current student makes TOPD on-policy for dLLMs. Unlike random masks, which can expose later answer tokens while hiding earlier reasoning tokens, $\tau$ follows the student's own reveal order and supplies the states encountered during inference.

\subsection{Trace-Aligned Decision Selection}
\label{subsec:trace_alignment}

A second challenge is that not every token proposal in a diffusion rollout is a final decision. Because later denoising may revise or overwrite provisional proposals, TOPD adopts the trace-construction principle of TraceRL \citep{wang2025tracerl}, where the trace keeps only token decisions that appear in the final response:
\begin{equation}
    \tilde{a}_t
    =
    \{(j,x_j)\in a_t : x_j=s_{T,j}\}.
\end{equation}
Here $j$ indexes a response position, and $\tilde{a}_t$ is the trace-aligned subset at diffusion step $t$.

Trace alignment turns a non-autoregressive rollout into supervised state-position pairs. Each retained pair $(s_t,j)$ marks the context in which the student committed final token $x_j$, so teacher feedback is attached to output-forming decisions rather than arbitrary masked positions in a completed answer.

\subsection{Step-Wise Teacher Distribution Matching}
\label{subsec:dense_supervision}

Teacher feedback must use the same conditional context in which the student made each retained decision. TOPD therefore evaluates teacher and student distributions on the same trace-aligned state-position pairs. Let $\pi_{\text{tea}}$ be a frozen teacher dLLM. For each $(j,x_j)\in\tilde{a}_t$, the teacher distribution is
\begin{equation}
    p^{\text{tea}}_{t,j}(\cdot)
    =
    \pi_{\text{tea}}(\cdot \mid q, s_t, j).
\end{equation}
The corresponding student distribution is evaluated at the same prompt, state, and position:
\begin{equation}
    p^{\theta}_{t,j}(\cdot)
    =
    \pi_{\theta}(\cdot \mid q, s_t, j).
\end{equation}
For block-wise implementations, $s_t$ denotes the schedule-consistent model input at that step.

This paired evaluation makes supervision state-matched rather than answer-matched. The teacher signal is dense at the token level, on-policy at the state level, and aligned with the student's diffusion trace.

\subsection{Reverse-KL Training Objective}
\label{subsec:objective}

The final challenge is to turn trace-aligned teacher distributions into stable token-level updates. TOPD matches each retained student distribution to the frozen teacher with Reverse-KL. For $B$ prompts and $G$ sampled trajectories per prompt, the loss is
\begin{equation}
\begin{aligned}
    \mathcal{L}_{\text{TOPD}}(\theta)
    &=
    \frac{1}{BG}
    \sum_{i=1}^{BG}
    \sum_{t:\,|\tilde{a}_{t,i}|>0}
    \frac{1}{|\tilde{a}_{t,i}|}
    \\
    &\quad
    \sum_{(j,x_j)\in\tilde{a}_{t,i}}
    D_{\mathrm{KL}}\!\left(
        p^{\theta}_{t,j}(\cdot)
        \,\middle\|\,
        p^{\text{tea}}_{t,j}(\cdot)
    \right).
\end{aligned}
\end{equation}
The normalization averages over sampled trajectories and then over retained positions within each diffusion step, preventing steps with many revealed tokens from dominating the update.

Reverse-KL concentrates the student on teacher-preferred modes at the same state. This is useful in noisy partially denoised contexts, where many alternatives are possible but only a few continue the student's reasoning path. The divergence ablation in Section~\ref{subsec:ablations} supports this choice.

In implementation, TOPD uses a sampled-token score-function estimator for the negative Reverse-KL objective. For each retained token $x_j$, define
\begin{equation}
    r_{t,j}
    =
    \mathrm{sg}\!\left(
        \log p^{\text{tea}}_{t,j}(x_j)
        -
        \log p^{\theta}_{t,j}(x_j)
    \right),
\end{equation}
where $\mathrm{sg}(\cdot)$ denotes stop-gradient, so the log-probability gap is treated as a scalar coefficient. The corresponding ascent estimator is
\begin{equation}
\begin{aligned}
    \nabla_{\theta}\mathcal{J}_{\text{TOPD}}(\theta)
    &\approx
    \frac{1}{BG}
    \sum_{i=1}^{BG}
    \sum_{t:\,|\tilde{a}_{t,i}|>0}
    \frac{1}{|\tilde{a}_{t,i}|}
    \\
    &\quad
    \sum_{(j,x_j)\in\tilde{a}_{t,i}}
    \nabla_{\theta}\log p^{\theta}_{t,j}(x_j)\, r_{t,j}.
\end{aligned}
\end{equation}
A retained token receives a positive update when the teacher assigns it higher probability than the student and a negative update otherwise. When full vocabularies are materialized, the same objective can be evaluated by summing over tokens. The sampled form avoids reward models, value heads, old-policy ratios, and sequence-level credit assignment.

%% file: 4-experiment.tex
\section{Experiments}

We evaluate TOPD along three axes: effectiveness, mechanism, and practical value. First, we test whether TOPD can match RL-based post-training on mathematical reasoning benchmarks under both static and dynamic decoding settings. Second, we isolate whether on-policy states, trace-aligned token selection, and Reverse-KL matching each contribute to the gain. Third, we measure whether dense teacher supervision reduces training cost without introducing training instability.

\subsection{Experimental Setup}
\label{subsec:experimental_setup}
Unless otherwise noted, experiments use SDAR-4B-Chat
\citep{cheng2025sdar} as the student and TraDo-8B-Instruct
\citep{wang2025tracerl} as the teacher. For mathematical post-training,
we use the MATH training set \citep{hendrycks2021measuring}, retaining
level 3--5 problems following \citet{hu2025open}, yielding 8K tasks. We
use block-wise diffusion with 4-token blocks, 4 denoising steps per block,
and 64 prompts per rollout round. A \emph{rollout step} denotes one
rollout round plus its associated training update, not a single optimizer
step.

Training rollouts use dynamic decoding. TOPD samples one response per
prompt, whereas TraceRL samples 16 for reward normalization and
value-based credit assignment. SFT uses TraDo-8B-Instruct-generated
responses from the same data source with the semi-autoregressive objective
for block-attention dLLMs \citep{arriola2025block}. We evaluate every 5
rollout steps and report avg@3 for MATH500
\citep{hendrycks2021measuring} and GSM8K \citep{cobbe2021training}, and
avg@20 for AIME2024 \citep{maa2024aime}, under both static and dynamic
decoding.

\subsection{Main Results}

\begin{table*}[t]
\centering
\small
\setlength{\tabcolsep}{4.2pt}
\caption{Main results on MATH500, AIME2024, and GSM8K. ``Avg.'' is the arithmetic mean over the three benchmarks for each decoding strategy. For TOPD, scores are reported from the best evaluation checkpoint. Method prefixes denote post-training applied to the base model; TraceRL-trained SDAR models correspond to TraDo checkpoints. ``Per-Round Speedup'' reports rollout-round model-compute speedup from Section~\ref{subsec:efficiency} and Appendix~\ref{app:training_cost}, with ESPO and TraceRL as the LLaDA and SDAR references (1$\times$). ``To-Accuracy Speedup'' further accounts for rollout rounds to the comparable MATH500 comparison point.}
\label{tab:main_results}
\resizebox{\textwidth}{!}{%
\begin{tabular}{lcccccccccc}
\toprule
\textbf{Model} &
\multicolumn{2}{c}{\textbf{MATH500}} &
\multicolumn{2}{c}{\textbf{AIME2024}} &
\multicolumn{2}{c}{\textbf{GSM8K}} &
\multicolumn{2}{c}{\textbf{Avg.}} &
\textbf{Per-Round} &
\textbf{To-Accuracy} \\
\cmidrule(lr){2-3}
\cmidrule(lr){4-5}
\cmidrule(lr){6-7}
\cmidrule(lr){8-9}
& \textbf{Static} & \textbf{Dynamic}
& \textbf{Static} & \textbf{Dynamic}
& \textbf{Static} & \textbf{Dynamic}
& \textbf{Static} & \textbf{Dynamic}
& \textbf{Speedup} & \textbf{Speedup} \\
\midrule
\textit{\scalebox{0.8}{Full Attention Model}}\\
LLaDA-8B-Instruct
& 37.3 & 38.3
& 0.5 & \textbf{1.7}
& 82.5 & 82.5
& 40.1 & 40.8
& -- & -- \\
\quad + ESPO
& 38.9\posdelta{1.6} & \textbf{39.1\posdelta{0.8}}
& \textbf{1.0\posdelta{0.5}} & 0.8\negdelta{0.9}
& \textbf{84.0\posdelta{1.5}} & \textbf{83.9\posdelta{1.4}}
& \textbf{41.3\posdelta{1.2}} & \textbf{41.3\posdelta{0.5}}
& 1$\times$ & 1$\times$ \\
\rowcolor[gray]{0.92}
\quad + TOPD
& \textbf{39.2\posdelta{1.9}} & 38.6\posdelta{0.3}
& 0.8\posdelta{0.3} & 1.2\negdelta{0.5}
& 83.3\posdelta{0.8} & 83.5\posdelta{1.0}
& 41.1\posdelta{1.0} & 41.1\posdelta{0.3}
& \textbf{10.9$\times$} & \textbf{511.6$\times$} \\
\midrule
\textit{\scalebox{0.8}{Block Attention Model}}\\
TraDo-8B-Instruct
& 78.5 & 75.5
& 13.3 & 11.0
& 92.3 & 91.2
& 61.4 & 59.2
& -- & -- \\
SDAR-4B-Chat
& 70.2 & 67.4
& 5.0 & 8.2
& 90.2 & 88.9
& 55.1 & 54.8
& -- & -- \\
\quad + SFT
& 74.3\posdelta{4.1} & 68.0\posdelta{0.6}
& 9.0\posdelta{4.0} & 7.3\negdelta{0.9}
& 91.4\posdelta{1.2} & 89.7\posdelta{0.8}
& 58.2\posdelta{3.1} & 55.0\posdelta{0.2}
& -- & -- \\
\quad + TraceRL
& 75.6\posdelta{5.4} & 71.8\posdelta{4.4}
& 8.3\posdelta{3.3} & \textbf{10.3\posdelta{2.1}}
& 91.2\posdelta{1.0} & \textbf{90.3\posdelta{1.4}}
& 58.4\posdelta{3.3} & \textbf{57.5\posdelta{2.7}}
& 1$\times$ & 1$\times$ \\
\rowcolor[gray]{0.92}
\quad + TOPD
& \textbf{75.9\posdelta{5.7}} & \textbf{71.9\posdelta{4.5}}
& \textbf{9.3\posdelta{4.3}} & 8.7\posdelta{0.5}
& \textbf{92.2\posdelta{2.0}} & 89.3\posdelta{0.4}
& \textbf{59.2\posdelta{4.1}} & 56.6\posdelta{1.8}
& \textbf{24.0$\times$} & \textbf{96.0$\times$} \\
\bottomrule
\end{tabular}
}
\end{table*}

\begin{figure}[t]
\centering
\includegraphics[width=\columnwidth]{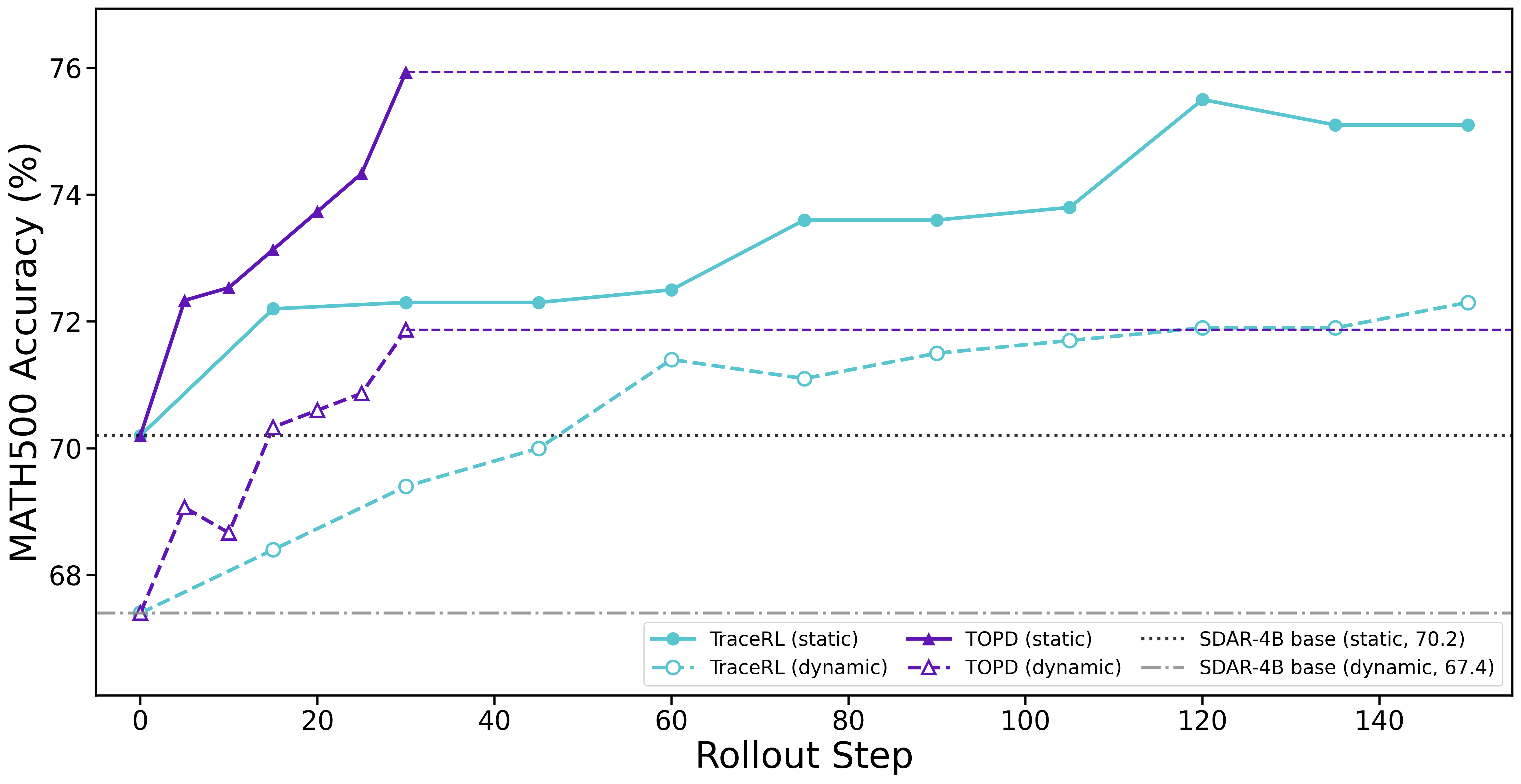}
\caption{Learning curves of TOPD vs.\ TraceRL on MATH500 under static (solid) and dynamic (dashed) masking. TOPD reaches strong performance by rollout step 30, while TraceRL requires 120 rollout steps for comparable performance.}
\label{fig:topd_vs_tracerl}
\end{figure}

TOPD matches the strongest SDAR-4B-Chat TraceRL gains on MATH500 while using far fewer rollouts. Table~\ref{tab:main_results} shows that TOPD improves SDAR-4B-Chat by +5.7 under static decoding and +4.5 under dynamic decoding, slightly exceeding TraceRL on MATH500 in both settings. SFT also improves the base model, but its gains are smaller than TOPD, especially under dynamic decoding. The improvements from TOPD extend to AIME2024 and GSM8K, showing that the transferred reasoning ability is not limited to the primary benchmark.

TOPD reaches strong performance substantially earlier than reward-based post-training. Figure~\ref{fig:topd_vs_tracerl} shows that TOPD reaches TraceRL-level MATH500 accuracy in roughly 4$\times$ fewer rollout rounds, supporting the claim that dense teacher supervision reduces the amount of sampled interaction needed for post-training.

\subsection{Generalization to Full-Attention Models}
\label{subsec:full_attention_generalization}

TOPD also transfers beyond the block-attention setting where it is primarily evaluated. In this experiment, the student is the original LLaDA-8B-Instruct \citep{nie2025largelanguagediffusionmodels}, a full-attention masked diffusion language model, and the teacher is the ESPO-trained LLaDA-8B-Instruct checkpoint \citep{ou2025espo}. As shown in Table~\ref{tab:main_results}, TOPD improves LLaDA-8B-Instruct on MATH500 and GSM8K, and its average gains are comparable to ESPO. The absolute gains are smaller than in the SDAR setting, likely because the LLaDA teacher is the student's RL-trained same-size counterpart rather than a larger model, but the positive trend suggests that trace-aligned teacher supervision is not specific to block-attention generation. Under the same teacher accounting, ESPO's reported policy-update budget and rollout grouping imply an estimated 10.9$\times$ per-round and 511.6$\times$ to-accuracy model-compute speedup for TOPD at a comparable MATH500 point (Appendix~\ref{app:training_cost}).

\subsection{Ablation Studies}
\label{subsec:ablations}

The ablation studies test whether each TOPD design choice is responsible for the observed gains. They follow the default SDAR setup in Section~\ref{subsec:experimental_setup} and evaluate each variant on MATH500 under both static and dynamic decoding. We vary only the source of supervised states, the token decisions that receive gradients, or the divergence objective.

For multi-run ablation summaries, we report \emph{peak mean accuracy}. At each evaluated training rollout step, we compute the mean accuracy and standard deviation across runs. We then choose the training rollout step whose across-run mean is highest and report that step's mean $\pm$ standard deviation. This statistic is therefore the peak of the mean learning curve.

\paragraph{On-policy vs.\ off-policy supervision.}

\begin{figure}[t]
\centering
\includegraphics[width=\columnwidth]{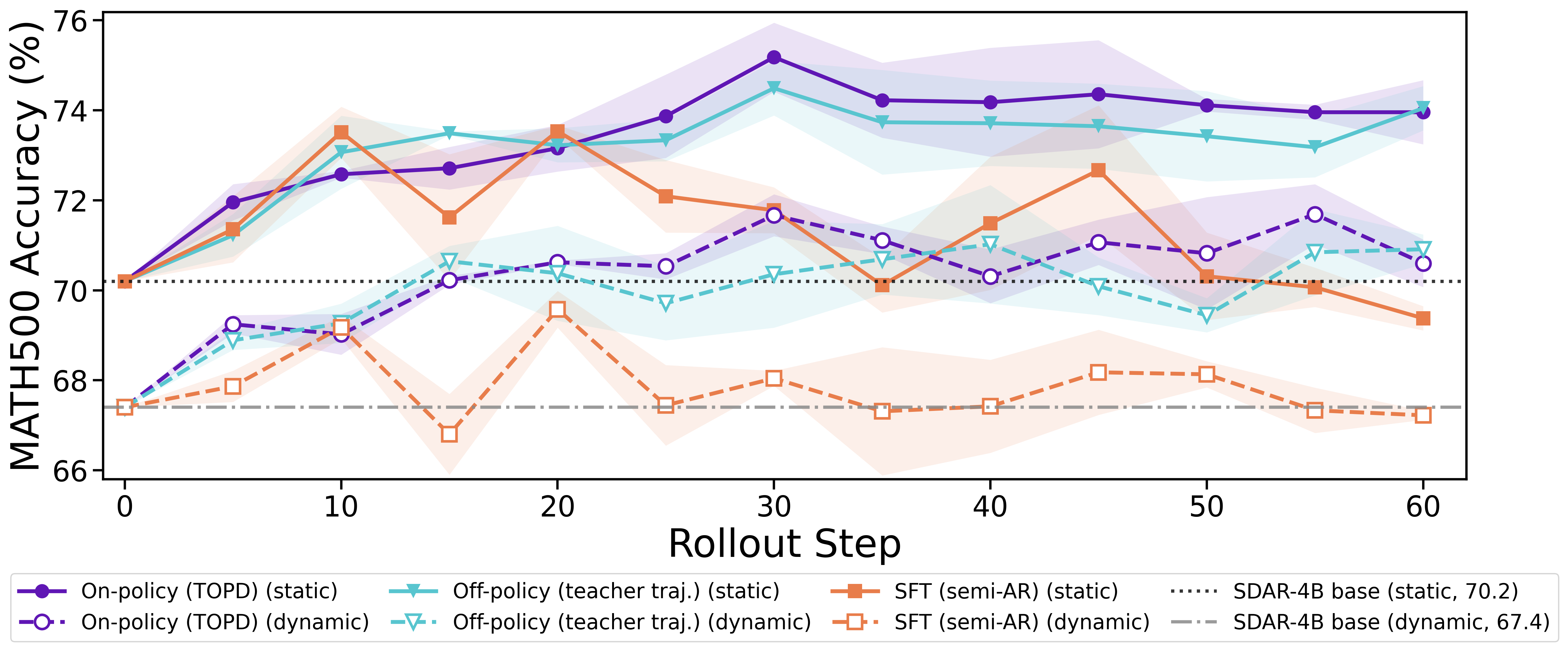}
\caption{On-policy, off-policy, and semi-AR SFT supervision on MATH500. Curves report mean accuracy over 3 runs with $\pm 1$ standard deviation shaded.}
\label{fig:ablation_on_off_policy}
\end{figure}

\begin{table}[t]
\centering
\small
\caption{On-policy, off-policy, and semi-AR SFT supervision on MATH500. Values are peak mean accuracy $\pm$ 1 standard deviation over 3 runs, as defined above; training rollout steps are shown in parentheses. On-policy and off-policy variants use the same Reverse-KL objective; semi-AR SFT follows the supervised setting in Section~\ref{subsec:experimental_setup}.}
\label{tab:ablation_on_off_policy}
\begin{tabular}{lcc}
\toprule
\textbf{Method} & \textbf{Static} & \textbf{Dynamic} \\
\midrule
Semi-AR SFT & 73.5$\pm$0.1 (20) & 69.6$\pm$0.4 (20) \\
Off-policy & 74.5$\pm$0.6 (30) & 71.0$\pm$1.3 (40) \\
On-policy (TOPD) & \textbf{75.2$\pm$0.8 (30)} & \textbf{71.7$\pm$0.7 (55)} \\
\bottomrule
\end{tabular}
\end{table}

Student-sampled states are more useful than fixed or purely supervised states. This ablation isolates the source of supervised states along an off-policy-to-on-policy spectrum. Semi-AR SFT uses fixed teacher-generated responses with a semi-autoregressive objective, off-policy distillation uses the same teacher-generated data as fixed trajectories, and TOPD samples trajectories from the current student. Thus, semi-AR SFT serves as the fixed-target supervised reference point, while off-policy distillation shares the Reverse-KL objective with TOPD but not its student-sampled states. Figure~\ref{fig:ablation_on_off_policy} shows the full across-run mean curves, and Table~\ref{tab:ablation_on_off_policy} shows an on-policy $>$ off-policy $>$ semi-AR SFT ranking under both decoding strategies.

\paragraph{Trace-based updates vs.\ random-mask supervision.}

\begin{figure}[t]
\centering
\includegraphics[width=\columnwidth]{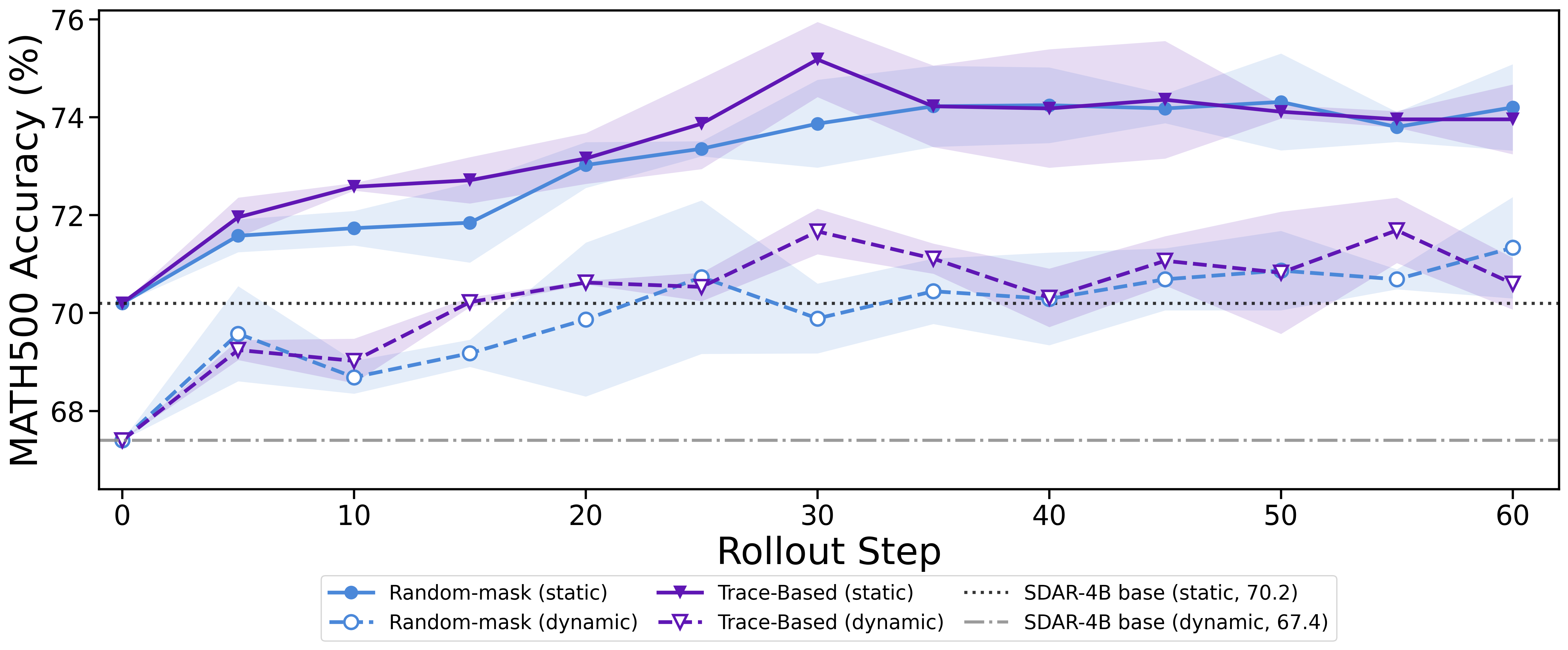}
\caption{Trace-based supervision vs.\ random-mask supervision on MATH500. Curves report mean accuracy over 3 runs with $\pm 1$ standard deviation shaded.}
\label{fig:ablation_trace_mc}
\end{figure}

\begin{table}[t]
\centering
\small
\caption{Trace-based versus random-mask supervision on MATH500. Values are peak mean accuracy $\pm$ 1 standard deviation over 3 runs, as defined above; training rollout steps are shown in parentheses.}
\label{tab:ablation_trace_mc}
\begin{tabular}{lcc}
\toprule
\textbf{Method} & \textbf{Static} & \textbf{Dynamic} \\
\midrule
Random-mask & 74.3$\pm$1.0 (50) & 71.3$\pm$1.0 (60) \\
Trace-aligned (TOPD) & \textbf{75.2$\pm$0.8 (30)} & \textbf{71.7$\pm$0.7 (55)} \\
\bottomrule
\end{tabular}
\end{table}

Trace-aligned token selection makes dense teacher feedback more targeted than random corruption. This ablation keeps the on-policy state distribution fixed and varies which denoising decisions receive teacher supervision. Random-mask supervision applies the teacher signal to randomly corrupted positions, whereas trace-aligned supervision restricts updates to token decisions that survive into the final response. Table~\ref{tab:ablation_trace_mc} shows that trace alignment improves peak mean accuracy under both decoding strategies, indicating that where the dense supervision is attached matters in addition to which states are sampled.

\paragraph{Divergence objectives: Reverse-KL, Forward Kullback--Leibler (Forward-KL), and Jensen--Shannon divergence (JSD).}

\begin{table}[t]
\centering
\small
\caption{Divergence objectives on MATH500. Single-run accuracy at key checkpoints. Step 30 and Step 60 are evaluated checkpoints, and $\Delta_{\mathrm{base}}$ is the peak improvement over SDAR-4B-Chat base accuracy (70.2 static, 67.4 dynamic). All variants use the same on-policy trace-based setting.}
\label{tab:kl_objective_math500}
\resizebox{\columnwidth}{!}{%
\begin{tabular}{lcccc}
\toprule
\textbf{Objective} & \textbf{Peak} & \textbf{Step 30} & \textbf{Step 60} & $\Delta_{\mathrm{base}} \uparrow$ \\
\midrule
\multicolumn{5}{c}{\textit{Static decoding}} \\
\midrule
Forward-KL & 72.7~(55) & 72.5 & 71.6 & +2.5 \\
JSD ($\beta = 0.5$) & 75.3~(35) & 74.5 & 73.6 & +5.1 \\
Reverse-KL (TOPD) & \textbf{75.9}~\textbf{(30)} & \textbf{75.9} & \textbf{73.8} & \textbf{+5.7} \\
\midrule
\multicolumn{5}{c}{\textit{Dynamic decoding}} \\
\midrule
Forward-KL & 68.9~(55) & 67.5 & 67.9 & +1.5 \\
JSD ($\beta = 0.5$) & 71.7~(60) & 69.5 & \textbf{71.7} & +4.3 \\
Reverse-KL (TOPD) & \textbf{72.6}~\textbf{(80)} & \textbf{71.9} & 70.8 & \textbf{+5.2} \\
\bottomrule
\end{tabular}
}
\end{table}

Reverse-KL is the most reliable divergence objective for trace-based distillation. This ablation keeps the on-policy, trace-aligned setting fixed and varies only the token-distribution matching objective. Table~\ref{tab:kl_objective_math500} shows that JSD is competitive, especially early in training, but Reverse-KL achieves the strongest peak overall. Forward-KL substantially underperforms, consistent with the analysis in Section~\ref{subsec:objective}: covering the teacher's full support in noisy states is less effective than concentrating on teacher-preferred modes.

Together, the ablations support the three core TOPD design choices. On-policy states improve over fixed targets, trace-aligned decisions improve over random masks, and Reverse-KL improves over Forward-KL and slightly over JSD at peak. These results explain why TOPD differs from SFT, fixed-trajectory distillation, and generic masked-token supervision.

\subsection{Efficiency and Stability}
\label{subsec:efficiency}

TOPD is more compute-efficient than RL because it removes group-rollout reward optimization. TraceRL samples 16 responses per prompt and requires old-logprob and value passes, whereas TOPD samples one response and performs a frozen-teacher forward plus one student update. Under our parameter-scaled accounting in Appendix~\ref{app:training_cost}, this yields a 24.0$\times$ per-round model-compute reduction for the SDAR-4B-Chat setting with value model.

This per-round saving compounds with faster convergence. TOPD reaches comparable MATH500 accuracy in 4$\times$ fewer rollout rounds, yielding an estimated to-accuracy speedup of 96.0$\times$ over TraceRL+V. For LLaDA, explicit rollout-round accounting gives a 10.9$\times$ per-round reduction; ESPO's 3K policy-update budget corresponds to 375 rollout steps, yielding a 511.6$\times$ speedup to the selected comparison checkpoint. We did not observe training collapse; the main late-training behavior is mild static overfitting after the peak checkpoint.

%% file: 5-relatedwork.tex
\section{Related Work}

\paragraph{Diffusion language models.}
Diffusion models generate by iterative denoising and have been adapted from continuous data to discrete language through categorical or masked corruption processes \citep{sohldickstein2015deep,ho2020ddpm,song2021score,austin2021d3pm,shi2024simplified}. Recent masked and block diffusion language models scale this paradigm to instruction following and reasoning with partially parallel generation \citep{nie2025largelanguagediffusionmodels,cheng2025sdar,arriola2025block,gong2025scaling,inception2025mercury,ye2025dream7bdiffusionlarge,bie2025llada20scalingdiffusionlanguage}. Because diffusion decoding produces provisional token decisions that may be revised before the final answer, trajectory structure matters \citep{wang2025tracerl,he2025mdpo}. TOPD therefore trains on the student's actual denoising trajectories rather than synthetic masked states.

\paragraph{Reinforcement learning for dLLM post-training.}
Reinforcement learning improves LLM reasoning through proximal policy optimization (PPO), reinforcement learning from human feedback (RLHF), preference optimization, and group-relative policy optimization \citep{schulman2015trpo,schulman2017ppo,ouyang2022instructgpt,rafailov2023dpo,shao2024deepseekmath,guo2025deepseek,yu2025dapo}. Recent work extends these ideas to dLLMs with masked SFT, trajectory-aware value modeling and sequence-level policy optimization \citep{zhao2025d1,wang2025tracerl,zhong2026stabledrl,ou2025espo,zhu2025llada,he2025mdpo,wang2025spg,zhao2025diffpo}. These methods show the value of on-policy rollouts and trace-aware trajectory construction, but sparse final rewards must be converted into useful step-level signals through likelihood ratios, value models, trajectory estimators, or stabilization heuristics. TOPD keeps on-policy trajectory coverage while replacing reward-based credit assignment with dense teacher supervision on the student's visited denoising states.

\paragraph{On-policy distillation.}
Knowledge distillation transfers behavior from a stronger teacher to a student by matching predictions or softened distributions \citep{hinton2015distillingknowledgeneuralnetwork,sanh2019distilbert,gu2024minillm,song2026survey}. Standard offline distillation trains on fixed teacher data and may not cover states caused by student errors, a classic exposure-bias issue in imitation learning \citep{ross2011reduction}. On-policy distillation instead samples states from the current student and queries teacher distributions on those states, combining on-policy coverage with dense low-variance supervision \citep{agarwal2024policy,lu2025onpolicydistillation,xuspeculative,song2026surveyonpolicydistillationlarge}.  TOPD extends this idea to diffusion generation, where supervision must be aligned with the final denoising trace rather than committed autoregressive prefixes.

%% file: 6-conclusion.tex
\section{Conclusion}

We introduced TOPD, a trace-based on-policy distillation framework for post-training masked diffusion language models. TOPD samples the student's own denoising trajectories, queries a frozen teacher on the corresponding partially denoised states, and applies Reverse-KL updates to the trace-aligned token decisions that form the final response. This preserves dense teacher supervision while avoiding sparse reward assignment, value modeling, and likelihood-ratio machinery. On mathematical reasoning, TOPD recovers TraceRL-level MATH500 gains for SDAR-4B-Chat with 4$\times$ fewer rollout rounds and an estimated 96.0$\times$ to-accuracy model-compute speedup; additional results on AIME2024, GSM8K, full-attention LLaDA, and ablations support the roles of on-policy states, trace alignment, and Reverse-KL matching. 

\section*{Limitations}
\label{app:limitations}

Although TOPD provides an efficient way to transfer reasoning ability from a stronger dLLM to a weaker student through dense on-policy teacher supervision, it still has several limitations. First, our experiments use an 8B dLLM as the teacher and a 4B dLLM as the student, because stronger and larger publicly available dLLM teachers are currently limited. For the full-attention experiments, the lack of model families with multiple parameter scales further requires using an RL-trained same-size model as the teacher. Therefore, it remains unclear whether the observed efficiency gains generalize when substantially larger or more capable teacher models become available. Second, the evaluation is limited to mathematical reasoning benchmarks, including MATH500, AIME2024, and GSM8K, and does not cover broader reasoning tasks such as code generation, tool use, or open-ended instruction following. Third, TOPD relies on a frozen teacher for token-level distributional supervision, making its effectiveness dependent on the teacher's capability and alignment with the target domain.

\section*{Acknowledgments}
The authors used AI assistants, including ChatGPT, for language polishing, wording suggestions, and coding assistance. All research ideas, experimental design, analyses, and final manuscript content were reviewed and verified by the authors.

%% file: 7-appendix.tex
\section{Additional Method Details}
\label{app:method_details}

This section formalizes the auxiliary definitions used by the method and ablation studies. We first define the attention patterns for full-attention and block-attention dLLMs, then specify the supervised, decoding, random-mask, and distribution-matching objectives referenced in the main text.

\subsection{Attention Patterns}
\label{app:attention_patterns}

Let a response have length $L$ and let $B$ denote the block size. We write $b(i)=\lceil i/B\rceil$ for the response block containing position $i$. Prompt tokens are visible to all response positions in both model families, so the distinction below concerns response-to-response attention. Following prior masked diffusion language models and block diffusion models \citep{sahoo2024simple,nie2025largelanguagediffusionmodels,arriola2025block,cheng2025sdar}, a full-attention dLLM uses a bidirectional response mask
\begin{equation}
    A^{\mathrm{full}}_{ij}=1,
    \qquad
    1\le i,j\le L,
    \label{eq:full_attention_mask}
\end{equation}
which allows every response position to condition on every other response position in the corrupted sequence. A block-attention dLLM uses a block-causal response mask
\begin{equation}
    A^{\mathrm{block}}_{ij}
    =
    \mathbf{1}\!\left[b(j)\le b(i)\right],
    \qquad
    1\le i,j\le L,
    \label{eq:block_attention_mask}
\end{equation}
so a token in block $b(i)$ can attend to tokens in the same block and earlier blocks, but not to future blocks. This is the architectural difference behind the ``full attention'' and ``block attention'' groupings in Table~\ref{tab:main_results}. In the SDAR setting, $B=4$.

\subsection{Masked and Semi-Autoregressive SFT}
\label{app:semi_ar_sft}

For a prompt-response pair $(q,y)$, where $y=(y_1,\ldots,y_L)$, the full-response masked SFT objective samples a corrupted response $\tilde{y}$ from $y$ and predicts the masked positions. For compactness, write $h^{\mathrm{full}}_j=(q,\tilde{y},j)$:
\begin{equation}
\begin{aligned}
    \mathcal{L}_{\mathrm{full\text{-}SFT}}(\theta)
    &=
    \mathbb{E}_{q,y,\tilde{y}}
    \frac{1}{|\mathcal{M}(\tilde{y})|}
    \Bigg[
    \\
    &\quad
    \sum_{j\in\mathcal{M}(\tilde{y})}
    -\log \pi_{\theta}(y_j\mid h^{\mathrm{full}}_j)
    \Bigg].
\end{aligned}
\label{eq:full_sft_appendix}
\end{equation}
where $\mathcal{M}(\tilde{y})=\{j:\tilde{y}_j=\texttt{[MASK]}\}$ is the masked-position set. This objective is appropriate for full-attention masked diffusion models because all response positions share one bidirectional corrupted context.

For block-attention dLLMs, the SFT baseline in Section~\ref{subsec:experimental_setup} uses a semi-autoregressive objective following block diffusion training \citep{arriola2025block}. Let
\begin{equation}
    C_k=\{(k-1)B+1,\ldots,\min(kB,L)\}
\end{equation}
be the $k$-th response block, and let $\tilde{y}_{C_k}$ be a corrupted copy of the current block. The clean prefix $y_{<C_k}$ is provided as context, while the loss is applied only inside the current block. Let $h^{\mathrm{semi}}_{k,j}=(q,y_{<C_k},\tilde{y}_{C_k},j)$:
\begin{equation}
\begin{aligned}
    \mathcal{L}_{\mathrm{semi\text{-}AR}}(\theta)
    &=
    \mathbb{E}_{q,y}
    \sum_{k=1}^{\lceil L/B\rceil}
    \mathbb{E}_{\tilde{y}_{C_k}}
    \frac{1}{|\mathcal{M}(\tilde{y}_{C_k})|}
    \Bigg[
    \\
    &\quad
    \sum_{j\in\mathcal{M}(\tilde{y}_{C_k})}
    -\log \pi_{\theta}
    \bigl(y_j\mid h^{\mathrm{semi}}_{k,j}\bigr)
    \Bigg].
\end{aligned}
\label{eq:semi_ar_sft_appendix}
\end{equation}
Equation~\ref{eq:semi_ar_sft_appendix} is the supervised baseline denoted ``Semi-AR SFT'' in the ablations. It differs from TOPD in two ways: the target responses are fixed teacher-generated samples, and the supervised states are produced by an SFT masking process rather than by the current student's rollout trajectory.

\subsection{Decoding Schedules}
\label{app:decoding_schedules}

At diffusion step $t$, let $M_t=\{j:s_{t,j}=\texttt{[MASK]}\}$ be the currently masked response positions, and let
\begin{equation}
    c_{t,j}=\max_{v\in\mathcal{V}}\pi_\theta(v\mid q,s_t,j)
\end{equation}
be the model confidence at position $j$. We use confidence-based remasking schedules following recent dLLM decoding and trajectory-optimization work \citep{he2025mdpo,wang2025tracerl}. Static decoding reveals a fixed number of positions per step, usually the highest-confidence positions needed to satisfy the block schedule:
\begin{equation}
    U_t^{\mathrm{static}}
    =
    \operatorname{TopK}_{j\in M_t}(c_{t,j};\,m_t),
    \label{eq:static_decoding_appendix}
\end{equation}
where $m_t$ is determined by the block size and remaining denoising steps. Dynamic decoding instead reveals positions whose confidence exceeds a threshold $\tau$:
\begin{equation}
    U_t^{\mathrm{dynamic}}
    =
    \{j\in M_t:c_{t,j}\ge \tau\},
    \label{eq:dynamic_decoding_appendix}
\end{equation}
with a fallback to at least one revealed position when the set is empty. Both schedules then sample or select token values for $j\in U_t$ and keep the remaining positions masked. These definitions correspond to the static and dynamic evaluation columns in Section~\ref{subsec:experimental_setup}.

\subsection{Random-Mask Supervision}
\label{app:random_mask_mc}

The random-mask ablation uses the same sampled final response $s_T$ as TOPD but changes where teacher supervision is attached. We estimate this objective with $M$ independently sampled random masks. For each sample $r\in\{1,\ldots,M\}$, a random mask set $R^{(r)}\subseteq\{1,\ldots,L\}$ is drawn and a corrupted response $\bar{s}^{(r)}$ is formed by replacing positions in $R^{(r)}$ with \texttt{[MASK]}. In contrast to Semi-AR SFT, which applies one random masking instance to each fixed teacher response during a training pass, random-mask distillation expands each sampled response with enough independently drawn masks to match the number of trace rows used by TOPD. This keeps the amount of distillation supervision comparable to TOPD while removing trace-aligned position selection. Let $\bar{h}^{(r)}_j=(q,\bar{s}^{(r)},j)$, $p^\theta_{r,j}=\pi_\theta(\cdot\mid\bar{h}^{(r)}_j)$, and $p^{\mathrm{tea}}_{r,j}=\pi_{\mathrm{tea}}(\cdot\mid\bar{h}^{(r)}_j)$. The teacher and student are then matched on those random positions:

\begin{equation}
\begin{aligned}
    \mathcal{L}_{\mathrm{Random\text{-}Mask}}(\theta)
    &=
    \frac{1}{M}
    \sum_{r=1}^{M}
    \frac{1}{|R^{(r)}|}
    \sum_{j\in R^{(r)}}
    \\
    &\quad
    D\!\left(
    p^\theta_{r,j}\,\middle\|\,p^{\mathrm{tea}}_{r,j}
    \right).
\end{aligned}
\label{eq:random_mc_appendix}
\end{equation}
Here $D$ is the chosen distribution-matching objective. Unlike TOPD, this objective does not use the trace-aligned retained set $\tilde{a}_t$ and therefore may supervise positions unrelated to the student's actual reveal decisions.

\subsection{Distribution-Matching Objectives}
\label{app:matching_objectives}

For a retained or masked state-position pair, let $p=p^\theta_{t,j}$ denote the student distribution and $q=p^{\mathrm{tea}}_{t,j}$ denote the teacher distribution. Distribution matching is the standard signal in knowledge distillation and on-policy distillation \citep{hinton2015distillingknowledgeneuralnetwork,agarwal2024policy,lu2025onpolicydistillation}. The three objectives in Section~\ref{subsec:ablations} are
\begin{align}
    D_{\mathrm{RKL}}(p,q)
    &=
    D_{\mathrm{KL}}(p\|q)
    \notag\\
    &=
    \sum_{v\in\mathcal{V}}p(v)\log\frac{p(v)}{q(v)},
    \label{eq:rkl_appendix}
    \\
    D_{\mathrm{FKL}}(p,q)
    &=
    D_{\mathrm{KL}}(q\|p)
    \notag\\
    &=
    \sum_{v\in\mathcal{V}}q(v)\log\frac{q(v)}{p(v)},
    \label{eq:fkl_appendix}
\end{align}
and a $\beta$-weighted Jensen--Shannon objective. For $0<\beta<1$, let $m_\beta=(1-\beta)p+\beta q$ and define
\begin{equation}
\begin{aligned}
    D_{\mathrm{JSD},\beta}(p,q)
    &=
    (1-\beta)D_{\mathrm{KL}}(p\|m_\beta)
    \\
    &\quad+
    \beta D_{\mathrm{KL}}(q\|m_\beta).
\end{aligned}
    \label{eq:jsd_appendix}
\end{equation}
The ablation with $\beta=0.5$ uses the symmetric JSD. The implementation also permits endpoint settings by dispatching $\beta=0$ to Forward-KL and $\beta=1$ to Reverse-KL. In the main experiments, SDAR TOPD uses full-vocabulary KL, while LLaDA TOPD uses top-$k{=}10$ token KL for the teacher--student distribution match; Table~\ref{tab:llada_topd_hyperparams} reports this LLaDA-specific truncation.

\section{Algorithm Details}
\label{app:algorithm}

Algorithm~\ref{alg:topd} instantiates the TOPD loop from Section~\ref{sec:methodology}. TOPD adopts the trace-construction principle of TraceRL~\citep{wang2025tracerl}: after a diffusion rollout is completed, only token decisions that survive into the final response are retained as trace decisions. Each TOPD rollout round has three stages: sample a minibatch of prompts and responses from the current student, convert the resulting trajectories into trace-aligned distillation rows, and optimize on minibatches of those rows with frozen-teacher supervision.

\begin{algorithm}[htbp]
\caption{TOPD: Trace-Based On-Policy Distillation}
\label{alg:topd}
\begin{algorithmic}[1]
\REQUIRE Student $\pi_\theta$, frozen teacher $\pi_{\text{tea}}$, prompt set $\mathcal{Q}$, rollout prompt batch size $N$, responses per prompt $G$, update minibatch size $m$, block size $B$, denoising steps per block $K$, maximum response length $L_{\max}$
\FOR{each rollout round}
    \STATE Sample a prompt minibatch $\mathcal{B}=\{q_i\}_{i=1}^{N} \sim \mathcal{Q}$
    \STATE Initialize trace-row dataset $\mathcal{D}_{\mathrm{TOPD}}\gets\varnothing$
    \STATE \textbf{Sample student rollouts}
    \FOR{each prompt $q_i\in\mathcal{B}$}
        \FOR{$g=1$ \TO $G$}
            \STATE $s_{0}^{i,g}\gets$ fully masked response of length $L_{\max}$
            \STATE $\tau_{i,g}\gets(s_{0}^{i,g})$
            \FOR{$t=0$ \TO $T-1$}
                \STATE $a_{t}^{i,g}\sim\pi_\theta(\cdot\mid q_i,s_{t}^{i,g})$
                \STATE $s_{t+1}^{i,g}\gets\text{unmask}(s_{t}^{i,g},a_{t}^{i,g})$
                \STATE Append $(a_{t}^{i,g},s_{t+1}^{i,g})$ to $\tau_{i,g}$
            \ENDFOR
        \ENDFOR
    \ENDFOR
    \STATE \textbf{Build trace-aligned training rows}
    \FOR{each trajectory $\tau_{i,g}=(s_0,a_0,\ldots,s_T)$}
        \FOR{$t=0$ \TO $T-1$}
            \STATE $\tilde{a}_{t}^{i,g}\gets\{(j,x_j)\in a_t:x_j=s_{T,j}\}$
            \FOR{each $(j,x_j)\in\tilde{a}_{t}^{i,g}$}
                \STATE Add row $(q_i,\hat{s}_{t}^{i,g},j,x_j)$ to $\mathcal{D}_{\mathrm{TOPD}}$
            \ENDFOR
        \ENDFOR
    \ENDFOR
    \STATE \textbf{Teacher-supervised minibatch updates}
    \FOR{each update minibatch $\mathcal{M}\subset\mathcal{D}_{\mathrm{TOPD}}$ with $|\mathcal{M}|=m$}
        \FOR{each row $(q,\hat{s},j,x_j)\in\mathcal{M}$}
            \STATE $p^{\text{tea}}_{q,\hat{s},j}\gets\pi_{\text{tea}}(\cdot\mid q,\hat{s},j)$
            \STATE $p^{\theta}_{q,\hat{s},j}\gets\pi_\theta(\cdot\mid q,\hat{s},j)$
            \STATE $\ell(q,\hat{s},j)\gets D_{\mathrm{KL}}\!\left(p^\theta_{q,\hat{s},j}\,\|\,p^{\text{tea}}_{q,\hat{s},j}\right)$
        \ENDFOR
        \STATE $\mathcal{L}_{\mathcal{M}}\gets\frac{1}{|\mathcal{M}|}\sum_{(q,\hat{s},j,x_j)\in\mathcal{M}}\ell(q,\hat{s},j)$
        \STATE $\theta\gets\theta-\eta\nabla_\theta\mathcal{L}_{\mathcal{M}}$
    \ENDFOR
\ENDFOR
\end{algorithmic}
\end{algorithm}

TOPD does not ask the teacher to supervise arbitrary corruptions of a completed answer; it supervises only the predictions that survive into $s_T$. Consequently, teacher feedback is attached to decisions that actually form the sampled response. The state $\hat{s}_t$ is reconstructed with the same schedule constraints as the student's denoising step: positions beyond the current progress point are reset to \texttt{[MASK]}, so teacher and student are evaluated under the same conditional context.

In practice, we implement Reverse-KL via a sampled-token score-function estimator:
\begin{equation}
    \nabla_\theta \mathcal{L}_{t,j} \approx
    -\nabla_\theta \log p^\theta_{t,j}(x_j)
    \cdot \mathrm{sg}\!\left(
        \log \tfrac{p^{\text{tea}}_{t,j}(x_j)}
                    {p^\theta_{t,j}(x_j)}
    \right).
\end{equation}
The estimator updates only the realized token $x_j$ and uses the teacher--student log-probability gap as a stopped scalar coefficient. If the teacher assigns higher probability to the retained token than the student, the update increases the student's probability of that token; if the teacher assigns lower probability, the update suppresses it. In implementation, teacher logits are computed once for each reconstructed trace state and reused for all retained positions at that step.

\section{Experimental Details}
\label{app:experimental_details}
\label{app:implementation}

This section collects the experimental choices and supplementary ablation evidence needed to reproduce and interpret the main TOPD runs. We separate default SDAR settings, prompt construction, the full-attention LLaDA variant, ablation settings, baseline settings, evaluation rules, and extended run-level analyses.

\subsection{Training Configuration}
\label{app:training_config}

Table~\ref{tab:topd_hyperparams} gives the default SDAR-4B-Chat TOPD configuration. Unless an ablation explicitly changes a component, the experiments in Section~\ref{subsec:ablations} inherit these model, rollout, optimization, and data settings.

\begin{table}[t]
\centering
\small
\caption{Training hyperparameters for TOPD on SDAR-4B-Chat.}
\label{tab:topd_hyperparams}
\resizebox{\columnwidth}{!}{%
\begin{tabular}{ll}
\toprule
\textbf{Category} & \textbf{Value} \\
\midrule
\multicolumn{2}{l}{\textit{Model \& Initialization}} \\
Student model & SDAR-4B-Chat\\
Teacher model & TraDo-8B-Instruct \\
Precision & bfloat16, TF32 enabled \\
Gradient checkpointing & Enabled (whole layer) \\
\midrule
\multicolumn{2}{l}{\textit{Rollout (On-Policy Sampling)}} \\
Prompts per round & 64 \\
Responses per prompt & 1 \\
Sampling temperature & 1.0 \\
Max generation tokens & 2000 \\
Block size / Denoising steps per block & 4 / 4 \\
Remasking strategy & Low-confidence dynamic ($\tau{=}0.9$) \\
Top-$p$ & 1.0 \\
\midrule
\multicolumn{2}{l}{\textit{Training \& Optimization}} \\
Optimizer & AdamW \\
Learning rate & $2 \times 10^{-7}$ (constant) \\
Weight decay & 0.0 \\
Effective batch size & 16 (4/GPU $\times$ 4 GPUs $\times$ 1 accum.) \\
Gradient clipping (max norm) & 1.0 \\
Training epochs per rollout & 1 \\
\midrule
\multicolumn{2}{l}{\textit{TOPD-Specific}} \\
Loss type & Full-vocab Reverse-KL\\
Teacher status & Frozen \\
Trace retention & Trace-aligned decisions only \\
Stop-gradient on log-prob gap & Enabled \\
\midrule
\multicolumn{2}{l}{\textit{Data}} \\
Training dataset & A subset of MATH (8,523 problems) \\
Max prompt length & 784 tokens \\
\bottomrule
\end{tabular}
}
\end{table}

\paragraph{Training infrastructure.}
All SDAR TOPD experiments use 4 NVIDIA A100-80G or A800 GPUs. We train with DeepSpeed ZeRO-2, parameter CPU offload, bfloat16 mixed precision, and TF32 enabled. A rollout round consists of student sampling followed by one training pass over the resulting traces; for SDAR-4B, one round takes approximately 8--12 minutes on 4 A100-80G GPUs, and the 30-step run used for the main comparison finishes in about 5 hours. For rollout and evaluation inference, we use an extended Fast-dLLM implementation \citep{wu2025fastdllm,wang2025tracerl} for LLaDA and JetEngine \citep{cheng2025sdar} for SDAR.

\subsection{Prompt Template and Rollout Records}
\label{app:prompt_template}

TOPD uses the same prompt surface for student rollouts, teacher supervision, and evaluation. The student first generates an on-policy denoising trace from the model-specific prompt below; the teacher is then queried on the reconstructed partially denoised states from that same trace. The teacher receives the original problem and the student's current state.
\begin{tcolorbox}[
    title={Math prompt templates},
    colback=gray!2,
    colframe=black!25,
    enhanced,
    fontupper=\small\ttfamily,
]
{\normalfont \textbf{SDAR}}

\begin{minipage}{\linewidth}
\raggedright
\textless|im\_start|\textgreater user\\
\{\{problem\}\}\\
Please reason step by step, and put your final answer within \textbackslash boxed\{\}.\textless|im\_end|\textgreater\\
\textless|im\_start|\textgreater assistant
\end{minipage}

{\normalfont \textbf{LLaDA}}

\begin{minipage}{\linewidth}
\raggedright
\seqsplit{\textless|startoftext|\textgreater\textless|start\_header\_id|\textgreater user\textless|end\_header\_id|\textgreater}
You need to put your final answer in \textbackslash boxed\{\}. This is the problem:\\
\{\{problem\}\}
\seqsplit{\textless|eot\_id|\textgreater\textless|startoftext|\textgreater\textless|start\_header\_id|\textgreater assistant\textless|end\_header\_id|\textgreater}
\end{minipage}
\end{tcolorbox}

Training uses the same problem statement and sampled response format as rollout generation for each model family. This keeps the training-time prompt identical to the rollout-time prompt and avoids introducing a separate reference-answer formatting path. Evaluation on MATH500, AIME2024, and GSM8K uses the corresponding model-specific template, with the benchmark problem substituted for \texttt{\{\{problem\}\}}. For MATH500 and AIME2024, the final \verb|\boxed{}| expression is used for answer extraction; for GSM8K, we also parse the final numeric answer to match the standard GSM8K evaluation convention.

\subsection{Full-Attention TOPD Configuration}
\label{app:llada_config}

Table~\ref{tab:llada_topd_hyperparams} specifies the full-attention LLaDA run used in Section~\ref{subsec:full_attention_generalization}. This experiment keeps the TOPD data flow fixed but changes the model family, generation budget, block size, and objective variant to match the LLaDA recipe and its ESPO comparison point.

\begin{table}[t]
\centering
\small
\caption{Training hyperparameters for TOPD on LLaDA-8B-Instruct.}
\label{tab:llada_topd_hyperparams}
\resizebox{\columnwidth}{!}{%
\begin{tabular}{ll}
\toprule
\textbf{Category} & \textbf{Value} \\
\midrule
Student model & LLaDA-8B-Instruct \\
Teacher model & LLaDA-8B-Instruct-ESPO\\
Training dataset & MATH training subset \\
Prompts / responses per rollout round & 32 / 1 \\
Training rollout rounds / selected checkpoint & 30 / 8 \\
Evaluation cadence & Every 2 rollout rounds \\
Rollout temperature & 0.8 \\
Max generation length / diffusion steps & 512 / 512 \\
Block size / further horizon & 32 / 128 \\
Remasking strategy & Low-confidence static \\
Batch size / gradient accumulation & 4 / 2 \\
Learning rate & $2 \times 10^{-7}$ (constant) \\
Loss type & Reverse-KL with top-$k{=}10$ token KL \\
Evaluation decoding & Static and dynamic, temperature 0.1 \\
\bottomrule
\end{tabular}
}
\end{table}

\subsection{Baseline Configurations}
\label{app:baseline_configs}

\paragraph{TraceRL.}
We implement TraceRL~\citep{wang2025tracerl} with the official training recipe. Each rollout round samples $G{=}16$ responses per prompt, scores completed responses with the reward model, recomputes old token log-probabilities, and applies a PPO-style update with clipping $\epsilon{=}0.2$ and KL penalty $\beta{=}0.01$. In the SDAR-4B comparison, TraceRL also uses a value model, which adds value inference and value optimization to each training cycle. The learning rate is $1 \times 10^{-6}$ with cosine decay, and the effective batch size is matched to TOPD at 16.

\paragraph{Semi-AR SFT.}
The supervised SDAR baseline fine-tunes SDAR-4B-Chat on the 8,523-problem MATH training subset using fixed teacher-generated responses. Following block diffusion training~\citep{arriola2025block}, it uses semi-autoregressive masking with block size $B{=}4$, a mask rate uniformly sampled from $[0.1,0.9]$, and cross-entropy on masked response tokens only. Training runs for one epoch with AdamW, learning rate $1\times10^{-5}$, effective batch size 16, and gradient clipping at 1.0. 

\paragraph{ESPO.}
We evaluate ESPO~\citep{ou2025espo} on LLaDA-8B-Instruct with LoRA adaptation on attention and MLP projection matrices ($r{=}128$, $\alpha{=}64$, dropout 0.05). Following the ESPO MATH setting, the policy update uses group size $G{=}16$, policy-update value $\mu{=}8$, $M{=}2$ Monte Carlo samples, bfloat16 precision, learning rate $1\times10^{-6}$, weight decay 0.1, and gradient clipping at 0.2. Generation uses random masking with 256 diffusion steps, maximum prompt length 400, maximum completion length 256, and temperature 1.0. TOPD-LLaDA uses the same prompt set and a 512-step decoding budget.

\subsection{Evaluation Protocol}
\label{app:eval_protocol}

We evaluate SDAR checkpoints under both static and dynamic decoding. Dynamic decoding uses a confidence threshold of $\tau=0.9$ and $top$-$k=0$, i.e., all tokens are retained. Static decoding follows Cheng et al.~\citep{cheng2025sdar} and uses $top$-$k=1$; both use block size 4 and 4 denoising steps per block. We sample with temperature 1.0 and top-$p$ 1.0, reporting avg@3 for MATH500 and GSM8K and avg@20 for AIME2024, where avg@$k$ is the average accuracy over $k$ sampled responses per problem. MATH500 contains 500 problems, AIME2024 contains 30 problems, and GSM8K uses the 1,319-example test split. For MATH500 and AIME2024, we extract the final \verb|\boxed{}| answer and compare it with SymPy-based normalization; for GSM8K, we extract the final numeric answer following the standard protocol.

\paragraph{Checkpoint selection.}
SDAR learning curves and ablations are evaluated every 5 rollout steps. Main-table accuracies are taken from the best single run for each method, using the checkpoint selected on MATH500 under the corresponding decoding setting and then evaluated on the remaining benchmarks. Multi-run ablation summaries report peak mean accuracy: at each evaluated step, we compute the across-run mean and standard deviation, then select the step with the highest across-run mean. Parenthesized step numbers in ablation tables denote the selected rollout step. Table~\ref{tab:main_checkpoint_summary} reports the MATH500 checkpoint choices for the main SDAR comparison and, for TOPD and TraceRL, the rollout-step contrast used in the compute-to-accuracy calculation.

\begin{table}[t]
\centering
\small
\caption{MATH500 checkpoint summary for the main SDAR comparison. Static and dynamic columns report avg@3 accuracy. ``Selection'' describes how the checkpoint is used in the main comparison.}
\label{tab:main_checkpoint_summary}
\resizebox{\columnwidth}{!}{%
\begin{tabular}{lccc}
\toprule
\textbf{Method} & \textbf{Selection} & \textbf{Static} & \textbf{Dynamic} \\
\midrule
SDAR-4B-Chat & Base checkpoint & 70.2 & 67.4 \\
Semi-AR SFT & Best single supervised checkpoint & 74.3 & 68.0 \\
TraceRL+V & Rollout step 120 comparison point & 75.5 & 71.9 \\
TOPD & Rollout step 30 comparison point & 75.9 & 71.9 \\
\bottomrule
\end{tabular}
}
\end{table}

\subsection{Response Length Analysis}
\label{app:response_length}

\begin{figure}[t]
\centering
\includegraphics[width=\columnwidth]{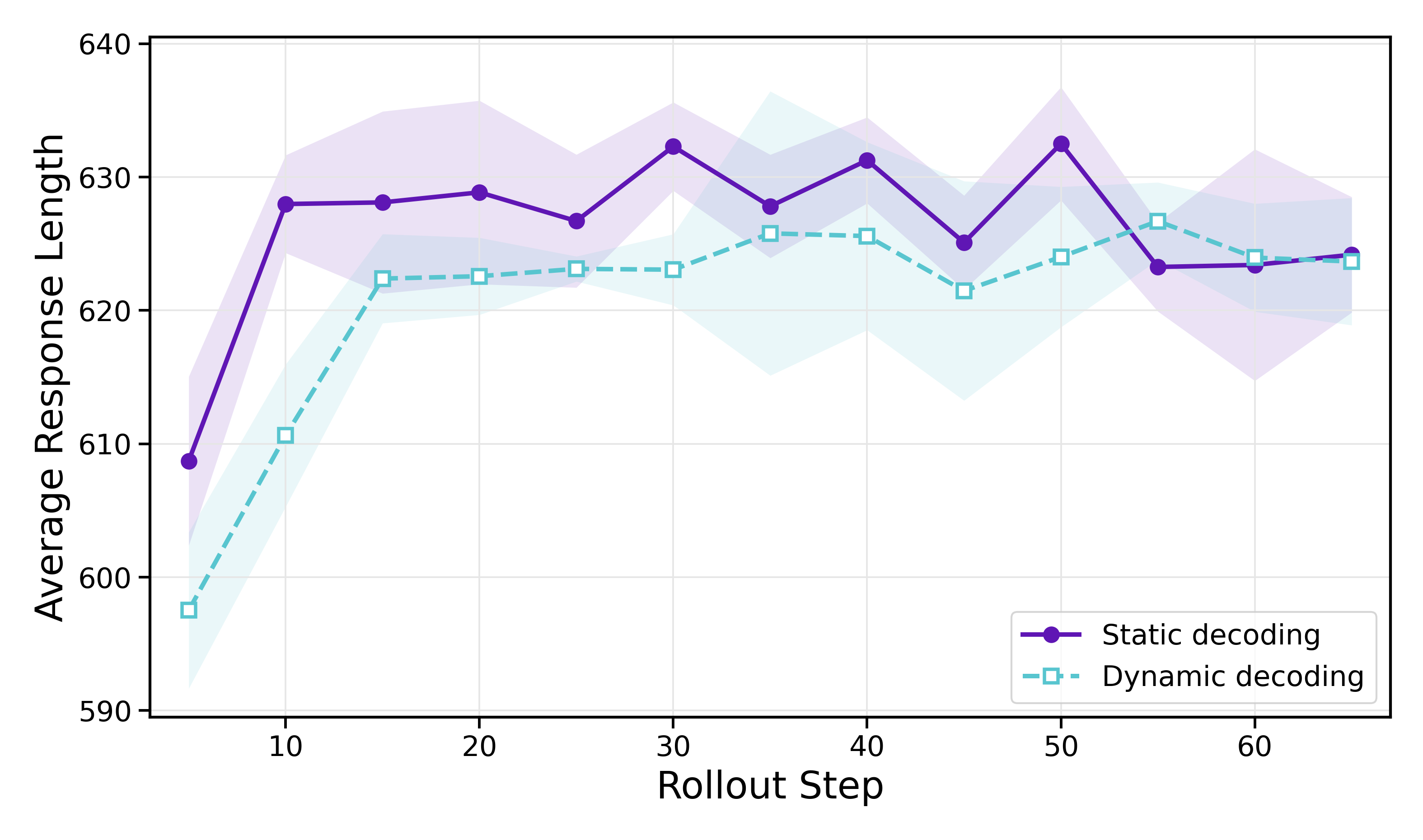}
\caption{Average response length on MATH500 for SDAR-4B-Chat + TOPD across three runs. Curves report the across-run mean, and shaded regions show $\pm 1$ standard deviation.}
\label{fig:response_length_math500}
\end{figure}

We additionally track generated response length for the SDAR-4B-Chat + TOPD runs to check whether TOPD's accuracy gains are accompanied by a large change in output budget. Figure~\ref{fig:response_length_math500} plots the MATH500 average response length over the three SDAR TOPD runs used in the on-policy ablation, with shaded regions showing $\pm 1$ standard deviation across runs. Response length rises during the first few rollout rounds and then stays in a narrow range under both static and dynamic decoding, suggesting that the main MATH500 improvements are not driven by an uncontrolled expansion of generated answers.

Table~\ref{tab:main_response_length} reports the average response length for the SDAR TOPD checkpoint used in Table~\ref{tab:main_results}.

\begin{table}[t]
\centering
\small
\caption{Average response length of SDAR-4B-Chat + TOPD corresponding to Table~\ref{tab:main_results}.}
\label{tab:main_response_length}
\resizebox{\columnwidth}{!}{%
\begin{tabular}{lcccc}
\toprule
\textbf{Benchmark} & \textbf{Static Acc.} & \textbf{Static Len.} & \textbf{Dynamic Acc.} & \textbf{Dynamic Len.} \\
\midrule
MATH500 & 75.9 & 628.8 & 71.9 & 620.2 \\
AIME2024 & 9.3 & 1151.5 & 8.7 & 1060.7 \\
GSM8K & 92.2 & 313.8 & 89.3 & 309.9 \\
\bottomrule
\end{tabular}
}
\end{table}

\subsection{Additional Ablation Diagnostics}
\label{app:extended_ablations}

This subsection provides an additional diagnostic view of the state-source ablation in Section~\ref{subsec:ablations}. All experiments use SDAR-4B-Chat as the student, TraDo-8B-Instruct as the frozen teacher, MATH500 as the benchmark, and the same training budget as the main SDAR setting.

\paragraph{Derived diagnostics for the state-source ablation.}
\label{app:state_source_diagnostics}

\begin{figure*}[h]
\centering
\includegraphics[width=\textwidth]{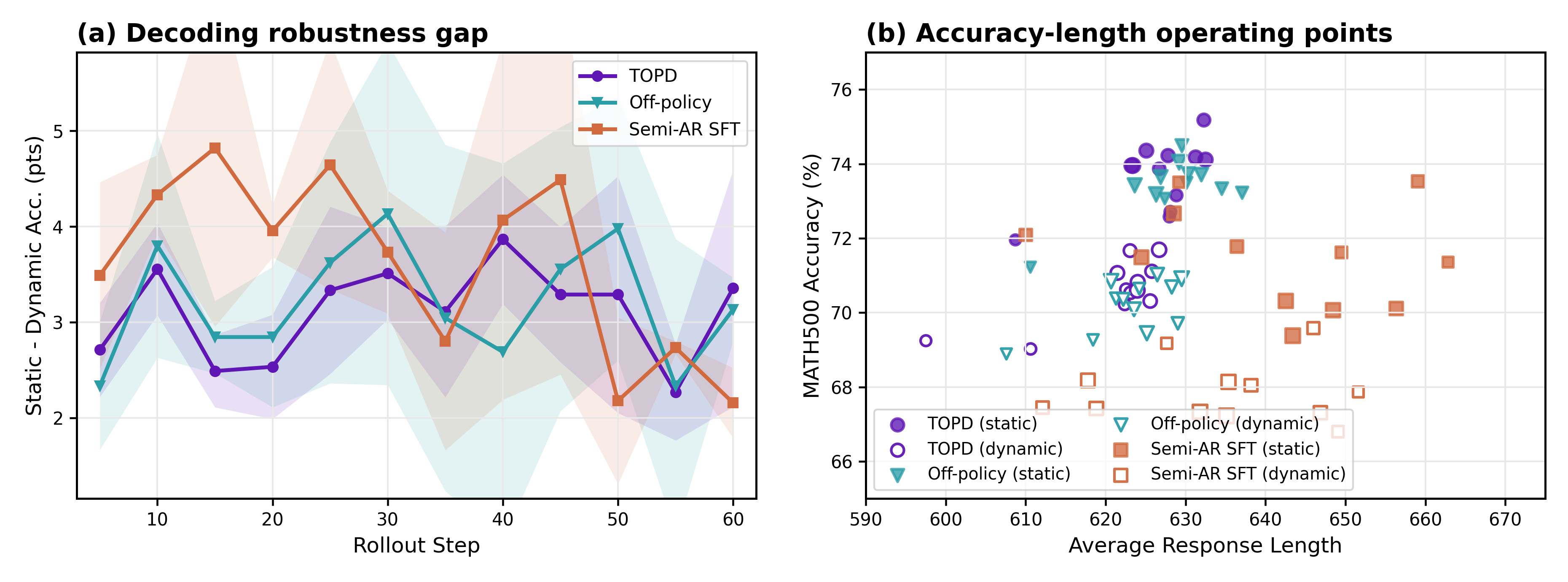}
\caption{Derived diagnostics for the MATH500 state-source ablation. Left: the static--dynamic accuracy gap over training, where smaller gaps indicate less sensitivity to the evaluation remasking schedule. Right: accuracy--length operating points for all evaluation steps with response-length logs; filled markers denote static decoding and open markers denote dynamic decoding.}
\label{fig:appendix_deep_ablation_analysis}
\end{figure*}

Figure~\ref{fig:appendix_deep_ablation_analysis} analyzes the same on-policy, off-policy, and Semi-AR SFT runs using two derived quantities rather than another copy of the learning curves. First, the static--dynamic gap shows that TOPD keeps the two evaluation schedules closer than Semi-AR SFT for most of training, indicating that trace-aligned on-policy supervision improves the sampled states. Second, the accuracy--length scatter shows that TOPD reaches the best operating region without moving to systematically longer responses: its high-accuracy static points cluster around the same response lengths as off-policy distillation, while Semi-AR SFT often produces longer responses without matching TOPD's dynamic accuracy. These diagnostics support the interpretation that TOPD's gains come from better state-matched supervision, not from a larger generation budget or a decoding-specific artifact.

\section{Case Study}
\label{app:qualitative}

We include three representative MATH500 generations from the SDAR-4B-Chat TOPD checkpoint at step~30 under static decoding. The first two are correct solutions that illustrate the model's ability to sustain multi-step, calculation-grounded reasoning over long trajectories. The third is a characteristic failure showing how a coherent-looking solution can go wrong through an early algebraic slip.

\tcbset{casebox/.style={
  colback=gray!2, colframe=black!30, coltitle=black,
  fonttitle=\bfseries\small, title=#1, before upper={\small}
}}

\tcbset{failbox/.style={
  colback=red!1.5!white, colframe=red!35!black, coltitle=white,
  fonttitle=\bfseries\small, title=#1, before upper={\small}, breakable
}}

\begin{tcolorbox}[casebox={Case 1: Coin Redistribution (MATH500~\#50)},breakable]

\textbf{Problem.} You have seven bags of gold coins. Each bag has the same number of gold coins. One day, you find a bag of 53 coins. You decide to redistribute the number of coins you have so that all eight bags you hold have the same number of coins. You successfully manage to redistribute all the coins, and you also note that you have more than 200 coins. What is the smallest number of coins you could have had before finding the bag of 53 coins?

\medskip
\textbf{TOPD output (correct, 203).}

Let the number of coins in each of the original seven bags be $x$, so the total is $7x$. After finding the bag of 53 coins, the total becomes $7x + 53$. After redistribution into eight equal bags, let each bag contain $y$ coins:

$$
8y = 7x + 53
$$

We also know the total exceeds 200: $8y > 200$, so $y > 25$. Since $y$ is integer, $y \ge 26$.

Substituting $y = 26$: $208 = 7x + 53 \Rightarrow 155 = 7x \Rightarrow x = 155/7 \not\in \mathbb{Z}$.

Continuing the search:
\begin{align*}
y=27&: 8\!\cdot\!27 = 7x\!+\!53 \;\Longrightarrow\; 163/7 \notin \mathbb{Z} \\
y=28&: 8\!\cdot\!28 = 7x\!+\!53 \;\Longrightarrow\; 171/7 \notin \mathbb{Z} \\
y=29&: 8\!\cdot\!29 = 7x\!+\!53 \;\Longrightarrow\; 179/7 \notin \mathbb{Z} \\
y=30&: 8\!\cdot\!30 = 7x\!+\!53 \;\Longrightarrow\; 187/7 \notin \mathbb{Z} \\
y=31&: 8\!\cdot\!31 = 7x\!+\!53 \;\Longrightarrow\; 195/7 \notin \mathbb{Z} \\
y=32&: 8\!\cdot\!32 = 7x\!+\!53 \;\Longrightarrow\; 203/7 = 29 \in \mathbb{Z}.
\end{align*}

The first integer solution is $x = 29$, giving the original total $7x = 203$. Hence $\boxed{203}$.

\end{tcolorbox}

\medskip
\begin{tcolorbox}[casebox={Case 2: Unfair Die Probability (MATH500~\#475)},breakable]

\textbf{Problem.} When rolling a certain unfair six-sided die with faces numbered 1, 2, 3, 4, 5, and 6, the probability of obtaining face $F$ is greater than $1/6$, the probability of obtaining the face opposite face $F$ is less than $1/6$, the probability of obtaining each of the other faces is $1/6$, and the sum of the numbers on each pair of opposite faces is 7. When two such dice are rolled, the probability of obtaining a sum of 7 is $\frac{47}{288}$. Given that the probability of obtaining face $F$ is $m/n$, where $m$ and $n$ are relatively prime positive integers, find $m+n$.

\medskip
\textbf{TOPD output (correct, 29).}

Let $p = P(F)$ and $q = P(\text{face opposite }F)$. The four remaining faces each have probability $1/6$. Since total probability sums to 1:

$$
p + q + 4 \cdot \frac{1}{6} = 1 \;\Longrightarrow\; p + q = \frac{1}{3}.
$$

The opposite-face pairs are $(1,6), (2,5), (3,4)$. Without loss of generality assume $F = 1$, so $P(1) = p$ and $P(6) = q$, while $P(2) = P(3) = P(4) = P(5) = 1/6$.

The probability of rolling sum 7 with two dice comes from six ordered pairs: $(1,6), (6,1), (2,5), (5,2), (3,4), (4,3)$. Therefore:

\begin{align*}
P(\text{sum}=7)
  &= 2pq + \tfrac{2}{36} + \tfrac{2}{36}
   = 2\bigl(pq + \tfrac{1}{18}\bigr).
\end{align*}

Given $P(\text{sum}=7) = 47/288$:
\begin{align*}
2\bigl(pq + \tfrac{1}{18}\bigr) &= \frac{47}{288} \\
\Rightarrow\; pq + \frac{1}{18} &= \frac{47}{576} \\
\Rightarrow\; pq &= \frac{47}{576} - \frac{32}{576}
= \frac{15}{576} = \frac{5}{192}.
\end{align*}

We now have the system $p + q = 1/3$ and $pq = 5/192$. Solving the quadratic $t^2 - \frac{1}{3}t + \frac{5}{192} = 0$:
\begin{align*}
192t^2 - 64t + 5 &= 0 \\[2pt]
t &= \frac{64 \pm \sqrt{64^2 - 4 \cdot 192 \cdot 5}}{2 \cdot 192} \\[2pt]
  &= \frac{64 \pm 16}{384} \\[2pt]
  &= \tfrac{5}{24} \text{ or } \tfrac{1}{8}.
\end{align*}

Since $p > 1/6$, we have $p = 5/24$ and $q = 1/8$. Thus $m = 5$, $n = 24$, and $m + n = 29$. Hence $\boxed{29}$.

\end{tcolorbox}

\medskip
\begin{tcolorbox}[failbox={Case 3 (Failure): Parallelogram Area (MATH500~\#239)}]

\textbf{Problem.} The parallelogram bounded by $y=ax+c$, $y=ax+d$, $y=bx+c$, $y=bx+d$ has area~18. The parallelogram bounded by $y=ax+c$, $y=ax-d$, $y=bx+c$, $y=bx-d$ has area~72. Given that $a,b,c,d$ are positive integers, find the smallest possible value of $a+b+c+d$.

\medskip
\textbf{TOPD output (incorrect;} predicts 20, ground truth is 16\textbf{).}

To solve the problem, we start by analyzing the given parallelograms and their areas. The first parallelogram is bounded by the lines $y = ax + c$, $y = ax + d$, $y = bx + c$, and $y = bx + d$. The second parallelogram is bounded by the lines $y = ax + c$, $y = ax - d$, $y = bx + c$, and $y = bx - d$.

\paragraph{Step 1: Determine the vertices.}
For the first parallelogram:
\begin{itemize}
\item $ax + c = bx + c \Rightarrow x = 0$, giving $(0, c)$.
\item $ax + d = bx + d \Rightarrow x = 0$, giving $(0, d)$.
\item $ax + c = bx + d \Rightarrow x = \frac{d-c}{a-b}$, yielding
  $\bigl(\frac{d-c}{a-b},\; \frac{ad-bc}{a-b}\bigr)$.
\item $ax + d = bx + c \Rightarrow x = \frac{c-d}{a-b}$, yielding
  $\bigl(\frac{c-d}{a-b},\; \frac{bc-ad}{a-b}\bigr)$.
\end{itemize}

For the second parallelogram:
\begin{itemize}
\item $ax + c = ax - d \Rightarrow x = -d/c$, giving $(-d/c,\; c-d)$.
\item $bx + c = bx - d \Rightarrow x = -d/c$, giving $(-d/c,\; c-d)$.
\item $ax + c = bx - d \Rightarrow x = \frac{c+d}{a-b}$, yielding
  $\bigl(\frac{c+d}{a-b},\; \frac{ac-bd}{a-b}\bigr)$.
\item $ax - d = bx + c \Rightarrow x = \frac{c+d}{a-b}$, yielding
  $\bigl(\frac{c+d}{a-b},\; \frac{bc-ad}{a-b}\bigr)$.
\end{itemize}

\paragraph{Step 2: Calculate the areas.}
For the first parallelogram, the distance between $y=ax+c$ and $y=ax+d$ is $|d-c|$, and between $y=bx+c$ and $y=bx+d$ is also $|d-c|$. The area is
\[
\frac{|d-c|}{\sqrt{1+a^2}} \cdot \frac{|d-c|}{\sqrt{1+b^2}}
= \frac{(d-c)^2}{\sqrt{(1+a^2)(1+b^2)}} = 18.
\]

For the second parallelogram, each distance is $2d$, giving
\[
\frac{2d}{\sqrt{1+a^2}} \cdot \frac{2d}{\sqrt{1+b^2}}
= \frac{4d^2}{\sqrt{(1+a^2)(1+b^2)}} = 72.
\]

\paragraph{Step 3: Solve.}
Divide the second equation by the first:
\begin{align*}
\frac{4d^2}{(d-c)^2} &= 4
\;\Longrightarrow\; d^2 = (d-c)^2 \\
\;\Longrightarrow\; d &= d-c \;\text{or}\; d = -(d-c).
\end{align*}

Since $d$ and $c$ are positive, $d = d-c$ is impossible, so $d = -(d-c) \Rightarrow 2d = c \Rightarrow c = 2d$.

Substituting $c = 2d$ back:
\[
\frac{d^2}{\sqrt{(1+a^2)(1+b^2)}} = 18.
\]

Squaring both sides:
\[
d^4 = 324\,(1+a^2)(1+b^2).
\]

Testing small values: $d=3$ gives $(1+a^2)(1+b^2)=1/4$, impossible. For $d=6$: $6^4=1296$, so $(1+a^2)(1+b^2)=4$, satisfied by $(a,b)=(1,1)$. With $c=2d=12$, we take $a=b=1$ and compute
$a+b+c+d = 1+1+12+6 = 20$.

Hence $\boxed{20}$.

\end{tcolorbox}

The failure case above (Case~3) illustrates the dominant error pattern we observed when inspecting incorrect TOPD generations: the model follows the requested step-by-step style, produces a boxed final answer, and the solution appears structurally sound---yet an early algebraic or arithmetic slip propagates unchecked through the derivation. Across the broader set of incorrect solutions, the most common failures are not formatting failures but \emph{local reasoning slips inside otherwise coherent solutions}: carrying an algebraic sign incorrectly, simplifying a fraction or exponent expression too early, mishandling a geometric constraint, or selecting a plausible intermediate result as the final answer before verifying it against the problem's exact requested quantity.

These failure modes are consistent with the aggregate results: TOPD improves the student's reasoning trajectory distribution, but it does not add an external verifier or symbolic correction mechanism. This observation motivates a natural extension: because TOPD already supplies dense teacher feedback on trace-aligned decisions, it could be combined with answer verification or process-level consistency checks---the teacher distribution would shape local denoising decisions, while a verifier would reject globally inconsistent solutions. We leave this combination to future work.

\section{Trace Mechanism: Progressive Denoising Visualization}
\label{app:trace_mechanism}

The following examples make trace-aligned supervision concrete. Each trajectory is drawn from MATH training data and decoded with block diffusion using block size $B{=}4$ and $K{=}4$ denoising steps per block. Within a block, positions are progressively revealed; positions that have not yet been produced are displayed as \mask{}. TOPD reconstructs these same partially denoised states for teacher evaluation, so the teacher supervises the student's actual reveal order rather than a randomly corrupted final answer.

\smallskip
\newcommand{\maskblock}{\mask{}\hspace{0.2em}\mask{}\hspace{0.2em}\mask{}\hspace{0.2em}$\cdots$\hspace{0.2em}\mask{}\hspace{0.2em}\mask{}\hspace{0.2em}\mask{}}
\newcommand{\maskshort}{\mask{}\,\mask{}\,\mask{}\,$\cdots$\,\mask{}\,\mask{}\,\mask{}}

\tcbset{tracebox/.style={
  colback=blue!0.5!white, colframe=black!25, coltitle=black,
  fonttitle=\bfseries\small, title=#1, before upper={\small}
}}

\subsection{Trace 1: Quadratic Equation (567~tokens, 48~blocks, 189~steps)}
\label{app:trace1}

\begin{tcolorbox}[tracebox={Trace 1: Within-Block Diffusion (Quadratic Equation)}]

\textbf{Problem.} Let $p$ and $q$ be the two distinct solutions to $(x-5)(2x+9) = x^2-13x+40$. What is $(p+3)(q+3)$?

\medskip
\noindent Block~1 produces ``\texttt{To solve the}'' over steps~1--4. Blocks~2--4 are shown below with one column per block position; \mask{} marks a position that remains masked at that denoising substep.\smallskip

{\footnotesize
\noindent
\begin{tabular}{r@{\hspace{0.2em}}c@{\hspace{0.5em}}c@{\hspace{0.5em}}c@{\hspace{0.5em}}c}
\multicolumn{5}{l}{\textit{Block 2 (steps 5--8):}} \\[2pt]
D$_1$ :& given   & \mask{} & \mask{} & \mask{} \\
D$_2$ :& given   & \mask{} & {\$}(    & \mask{} \\
D$_3$ :& given   & equation& {\$}(    & \mask{} \\
D$_4$ :& given   & equation& {\$}(    & x-5)( \\[6pt]

\multicolumn{5}{l}{\textit{Block 3 (steps 9--12):}} \\[2pt]
D$_1$ :& 2x+9   & \mask{} & \mask{} & \mask{} \\
D$_2$ :& 2x+9   & $) = x^{}$ & \mask{} & \mask{} \\
D$_3$ :& 2x+9   & $) = x^{}$ & 2-1    & \mask{} \\
D$_4$ :& 2x+9   & $) = x^{}$ & 2-1    & 3 \\[6pt]

\multicolumn{5}{l}{\textit{Block 4 (steps 13--16):}} \\[2pt]
D$_1$ :& x+40   & \mask{} & \mask{} & \mask{} \\
D$_2$ :& x+40   & {\$},    & \mask{} & \mask{} \\
D$_3$ :& x+40   & {\$},    & we     & \mask{} \\
D$_4$ :& x+40   & {\$},    & we     & first \\
\end{tabular}
}

\end{tcolorbox}

\medskip
\subsection{Trace 2: Diophantine Equation (880~tokens, 104~blocks, 413~steps)}
\label{app:trace2}

\begin{tcolorbox}[tracebox={Trace 2: Within-Block Diffusion (Diophantine Equation)}]

\textbf{Problem.} Find the number of ordered triples $(x,y,z)$ of real numbers such that $x^4 + y^4 + z^4 - 4xyz = -1$.

\medskip
\noindent Block~1 produces ``\texttt{To solve the equation}'' over steps~1--4. Blocks~2--4 use the same 4-column format, showing that even within a short prefix the reveal order can differ across positions.\smallskip

{\footnotesize
\noindent
\begin{tabular}{r@{\hspace{0.2em}}c@{\hspace{0.5em}}c@{\hspace{0.5em}}c@{\hspace{0.5em}}c}
\multicolumn{5}{l}{\textit{Block 2 (steps 5--8):}} \\[2pt]
D$_1$ :& {\$}    & \mask{}  & \mask{}  & \mask{} \\
D$_2$ :& {\$}    & $x^4+{}$ & \mask{}  & \mask{} \\
D$_3$ :& {\$}    & $x^4+{}$ & $y^4+{}$ & \mask{} \\
D$_4$ :& {\$}    & $x^4+{}$ & $y^4+{}$ & $z^4-{}$ \\[6pt]

\multicolumn{5}{l}{\textit{Block 3 (steps 9--12):}} \\[2pt]
D$_1$ :& 4xyz=  & \mask{}  & \mask{} & \mask{} \\
D$_2$ :& 4xyz=  & $-1$     & \mask{} & \mask{} \\
D$_3$ :& 4xyz=  & $-1$     & {\$}    & \mask{} \\
D$_4$ :& 4xyz=  & $-1$     & {\$}    & for \\[6pt]

\multicolumn{5}{l}{\textit{Block 4 (steps 13--16):}} \\[2pt]
D$_1$ :& \mask{}& \mask{}  & of      & \mask{} \\
D$_2$ :& \mask{}& number   & of      & \mask{} \\
D$_3$ :& the    & number   & of      & ordered \\
D$_4$ :& the    & number   & of      & ordered triples {\$}(x, \\
\end{tabular}
}

\end{tcolorbox}

\section{Training Compute Accounting}
\label{app:training_cost}

This appendix gives the update-normalized model-compute accounting used in
Section~\ref{subsec:efficiency}. The goal is not to predict wall-clock
runtime, but to compare the dominant neural-model forward and backward
work required by TOPD and the RL baselines under the same rollout-round
abstraction. We assume matched batch size, matched prompt budget, and the
same effective trace-row length across methods, so common multiplicative
factors are omitted.

\paragraph{FLOPs proxy.}
For a student model with $N_s$ parameters and effective trace-row length
$D$, we use
\begin{equation}
    C_{\mathrm{fwd}} = 2N_sD,\qquad
    C_{\mathrm{bwd}} = 4N_sD.
\end{equation}
Thus one trainable forward--backward update costs $6N_sD$. Let
$F_s=2N_sD$ denote one student forward-equivalent unit. A trainable
student update costs $3F_s$, and a frozen teacher forward costs
$\alpha F_s$, where $\alpha=N_t/N_s$. For the SDAR-4B experiments with an
8B teacher, $\alpha\approx2$; for same-size teacher accounting,
$\alpha=1$.

We use a full-model forward/backward proxy for all methods, including
LoRA-based baselines. This keeps the comparison at the level of dominant
model evaluations rather than trainable-parameter count or
implementation-specific optimizer savings.

\paragraph{Update-normalized costs.}
In the update-normalized proxy, response generation is counted as one
student forward-equivalent unit per sampled response. This abstracts away
the identical denoising schedule details and isolates the algorithmic
difference in the number of sampled responses and update-side model
passes.

TOPD samples one response per prompt, queries the frozen teacher once, and
performs one student update:
\begin{equation}
    C_{\mathrm{TOPD}}
    =
    F_s+\alpha F_s+3F_s
    =
    (4+\alpha)F_s.
\end{equation}

TraceRL samples $G=16$ responses per prompt. Without a value model, each
response requires response generation, old-log-probability recomputation,
and one policy update:
\begin{equation}
    C_{\mathrm{RL\text{-}noV}}
    =
    G(F_s+F_s+3F_s)
    =
    80F_s.
\end{equation}
Since the original TraceRL recipe uses a value model, we use the
value-model setting as the main comparison. The no-value path is reported
only as an auxiliary compute reference. With a value model trained every
$k$ rollout rounds, TraceRL additionally uses value inference and an
amortized value update:
\begin{equation}
\begin{aligned}
    C_{\mathrm{RL\text{-}V}}(k)
    &=
    G\left(
        F_s+F_s+F_s+3F_s+\frac{3F_s}{k}
    \right) \\
    &=
    16\left(6+\frac{3}{k}\right)F_s .
\end{aligned}
\end{equation}
For the SDAR-4B comparison, the value model is trained every round
($k=1$), giving
\begin{equation}
    C_{\mathrm{RL\text{-}V}}(1)=144F_s.
\end{equation}

Therefore the per-round update-normalized reductions are
\begin{align}
    S_{\mathrm{round,noV}}(\alpha)
    &=
    \frac{80}{4+\alpha},\\
    S_{\mathrm{round,V}}(\alpha,k)
    &=
    \frac{16(6+3/k)}{4+\alpha}.
\end{align}
For the SDAR-4B comparison with $\alpha=2$ and $k=1$, this gives
\begin{equation}
    S_{\mathrm{round,V}}(2,1)
    =
    \frac{144}{6}
    =
    24.0\times.
\end{equation}
With same-size teacher accounting, it gives $28.8\times$.

\begin{table}[t]
\centering
\small
\caption{Update-normalized per-round model-compute accounting for TraceRL
versus TOPD. $\alpha=1$ counts the teacher as one same-size frozen forward;
$\alpha=2$ charges the 8B teacher as twice the 4B student forward cost.}
\label{tab:compute_accounting}
\resizebox{\columnwidth}{!}{%
\begin{tabular}{lccc}
\toprule
\textbf{TraceRL setting} & \textbf{TraceRL cost} & $\alpha=1$ & $\alpha=2$ \\
\midrule
No value model & $80F_s$ & $16.0\times$ & $13.3\times$ \\
Value model, $k=2$ & $120F_s$ & $24.0\times$ & $20.0\times$ \\
Value model, $k=1$ & $144F_s$ & $28.8\times$ & $24.0\times$ \\
\bottomrule
\end{tabular}
}
\end{table}

\paragraph{Compute to reach comparable MATH500 accuracy.}
The learning-curve comparison compounds the per-round reduction with the
number of rollout rounds needed to reach comparable MATH500 accuracy. TOPD
reaches the comparison point at step 30, while TraceRL reaches it at step
120, giving a $4.0\times$ step ratio. For the value-model setting used in
the SDAR-4B comparison,
\begin{equation}
    24.0 \times 4.0 = 96.0\times .
\end{equation}
Under same-size teacher accounting, the corresponding estimate is
$115.2\times$; against the no-value TraceRL path, it is $53.3\times$.

\begin{table}[t]
\centering
\small
\caption{Estimated training speedups to reach comparable MATH500 accuracy
under parameter-scaled teacher accounting for the SDAR-4B comparison
($\alpha=2$, $k=1$).}
\label{tab:compute_to_accuracy}
\resizebox{\columnwidth}{!}{%
\begin{tabular}{lcc}
\toprule
\textbf{Comparison point} & \textbf{Step ratio} & \textbf{Speedup} \\
\midrule
Static, TOPD step 30 vs.\ TraceRL step 120 & $4.0\times$ & $96.0\times$ \\
Dynamic, TOPD step 30 vs.\ TraceRL step 120 & $4.0\times$ & $96.0\times$ \\
\bottomrule
\end{tabular}
}
\end{table}

\paragraph{ESPO versus TOPD on LLaDA.}
For the LLaDA comparison, we use a stricter rollout-round accounting so
that ESPO and TOPD are compared under the same data-collection unit. ESPO
reports a coupled-sampling FLOPs proxy in which one sampled response costs
\begin{equation}
    (K_{\mathrm{ESPO}} + 6\mu M)F_s,
\end{equation}
where $K_{\mathrm{ESPO}}$ is the number of diffusion sampling steps,
$\mu$ is the number of policy updates per data-collection round, and $M$
is the number of Monte Carlo ELBO samples. For the ESPO MATH setting,
$G=16$, $K_{\mathrm{ESPO}}=256$, $\mu=8$, and $M=2$, giving
\begin{align}
    C_{\mathrm{ESPO}}
    &=
    G(K_{\mathrm{ESPO}} + 6\mu M)F_s
    \notag\\
    &=
    16(256 + 6\times8\times2)F_s
    =
    5632F_s .
\end{align}
TOPD-LLaDA samples one response per prompt with $K_{\mathrm{TOPD}}=512$
diffusion steps, queries a same-size frozen teacher, and performs one
student update:
\begin{align}
    C_{\mathrm{TOPD}}
    &=
    (K_{\mathrm{TOPD}}+\alpha+3)F_s
    \notag\\
    &=
    (512+1+3)F_s
    =
    516F_s .
\end{align}
The resulting rollout-round model-compute speedup is
\begin{equation}
    S_{\mathrm{round,ESPO/TOPD}}
    =
    \frac{5632}{516}
    =
    10.9\times.
\end{equation}

\begin{table}[t]
\centering
\small
\caption{Rollout-round model-compute accounting for LLaDA ESPO and
TOPD. ESPO uses $G=16$, $K_{\mathrm{ESPO}}=256$, $\mu=8$, and $M=2$;
TOPD-LLaDA uses $K_{\mathrm{TOPD}}=512$ and a same-size frozen teacher.}
\label{tab:espo_topd_compute}
\resizebox{\columnwidth}{!}{%
\begin{tabular}{lcc}
\toprule
\textbf{Method} & \textbf{Dominant terms} & \textbf{Cost} \\
\midrule
ESPO-LLaDA & $G(K_{\mathrm{ESPO}}+6\mu M)F_s$ & $5632F_s$ \\
TOPD-LLaDA & rollout + teacher forward + student update & $516F_s$ \\
\bottomrule
\end{tabular}
}
\end{table}
For the LLaDA comparison, ESPO is trained for 3K policy-update steps under
its official recipe. Since ESPO uses $\mu=8$ policy updates per
data-collection round, this corresponds to $3000/8=375$ rollout rounds.
TOPD-LLaDA selects the step-8 checkpoint from a 30-round training run,
giving a rollout-step ratio of $375/8=46.875$. Combining this with the
unrounded per-round ratio gives
\begin{equation}
    \frac{5632}{516} \times 46.875 = 511.6\times .
\end{equation}
We therefore report $10.9\times$ per-round and $511.6\times$
to-accuracy model-compute speedups for the LLaDA comparison. This
same-size teacher accounting differs from the SDAR setting, where the
teacher is larger than the student; for LLaDA, the teacher is the
student's ESPO-trained counterpart.
\paragraph{Interpretation.}
These estimates are theoretical model-compute comparisons, not measured
wall-clock speedups. They intentionally abstract away implementation
details such as cached generation, synchronization, model placement,
rollout filtering, and dataloader overhead. The measured GPU-hour results
in Section~\ref{app:measured_gpu_hours} therefore provide a complementary
end-to-end runtime view.

\paragraph{Measured GPU-hour comparison and gap analysis.}
\label{app:measured_gpu_hours}

Table~\ref{tab:measured_gpu_hours} reports empirical GPU-hour measurements
from the SDAR-4B comparison on 4$\times$A100-80G GPUs. TraceRL with a
value model ($G{=}16$, $k{=}1$) consumes approximately 1.9--2.0 GPU-hours
per rollout round, while TOPD consumes 0.53--0.80 GPU-hours per round.
Combined with the faster convergence of TOPD, 30 versus 120 rollout
rounds to reach comparable MATH500 accuracy, this yields a measured
end-to-end speedup of $10$--$15\times$.

\begin{table}[t]
\centering
\small
\caption{Measured GPU-hour comparison between TOPD and TraceRL+V on
SDAR-4B using 4$\times$A100-80G GPUs. ``To comparable acc.'' denotes
GPU-hours to reach the accuracy points in Section~\ref{subsec:efficiency}.}
\label{tab:measured_gpu_hours}
\resizebox{\columnwidth}{!}{%
\begin{tabular}{lcc}
\toprule
\textbf{Metric} & \textbf{TraceRL+V} & \textbf{TOPD} \\
\midrule
GPU-hours per rollout round & 1.9--2.0 & 0.53--0.80 \\
GPU-hours per 10 rounds & 19--20 & 5.3--8.0 \\
Wall clock per round (4 GPUs) & $\sim$29 min & $\sim$8--12 min \\
GPU-hours to comparable static acc.\ (75.5--75.9\%) & 228--240 & 16--24 \\
GPU-hours to comparable dynamic acc.\ (71.9\%) & 228--240 & 16--24 \\
\bottomrule
\end{tabular}
}
\end{table}

The measured speedup is lower than the theoretical to-accuracy estimate
of $96.0\times$ because the model-compute proxy abstracts away
implementation-dependent costs. In practice, cached denoising forwards,
full-sequence training forwards, model placement, synchronization, and
pipeline overhead have different runtime profiles. TraceRL also updates
only on prompt groups that contain both correct and incorrect responses,
so not every sampled prompt contributes a policy-gradient update. These
factors reduce the realized per-round ratio, but the measured GPU-hour results still show
that TOPD reaches comparable MATH500 accuracy with substantially less
end-to-end training compute.

%% file: custom.bib
@InProceedings{sohldickstein2015deep,
  title = 	 {Deep Unsupervised Learning using Nonequilibrium Thermodynamics},
  author = 	 {Sohl-Dickstein, Jascha and Weiss, Eric and Maheswaranathan, Niru and Ganguli, Surya},
  booktitle = 	 {Proceedings of the 32nd International Conference on Machine Learning},
  pages = 	 {2256--2265},
  year = 	 {2015},
  editor = 	 {Bach, Francis and Blei, David},
  volume = 	 {37},
  series = 	 {Proceedings of Machine Learning Research},
  address = 	 {Lille, France},
  month = 	 {07--09 Jul},
  publisher =    {PMLR},
  url = 	 {https://proceedings.mlr.press/v37/sohl-dickstein15.html}
}

@inproceedings{ho2020ddpm,
  author       = {Jonathan Ho and
                  Ajay Jain and
                  Pieter Abbeel},
  editor       = {Hugo Larochelle and
                  Marc'Aurelio Ranzato and
                  Raia Hadsell and
                  Maria{-}Florina Balcan and
                  Hsuan{-}Tien Lin},
  title        = {Denoising Diffusion Probabilistic Models},
  booktitle    = {Advances in Neural Information Processing Systems 33: Annual Conference
                  on Neural Information Processing Systems 2020, NeurIPS 2020, December
                  6-12, 2020, virtual},
  year         = {2020},
  url          = {https://proceedings.neurips.cc/paper/2020/hash/4c5bcfec8584af0d967f1ab10179ca4b-Abstract.html},
  bibsource    = {dblp computer science bibliography, https://dblp.org}
}

@inproceedings{
song2021score,
title={Score-Based Generative Modeling through Stochastic Differential Equations},
author={Yang Song and Jascha Sohl-Dickstein and Diederik P Kingma and Abhishek Kumar and Stefano Ermon and Ben Poole},
booktitle={International Conference on Learning Representations},
year={2021},
url={https://openreview.net/forum?id=PxTIG12RRHS}
}

@inproceedings{
austin2021d3pm,
title={Structured Denoising Diffusion Models in Discrete State-Spaces},
author={Jacob Austin and Daniel D. Johnson and Jonathan Ho and Daniel Tarlow and Rianne van den Berg},
booktitle={Advances in Neural Information Processing Systems},
editor={A. Beygelzimer and Y. Dauphin and P. Liang and J. Wortman Vaughan},
year={2021},
url={https://openreview.net/forum?id=h7-XixPCAL}
}

@inproceedings{
li2022diffusionlm,
title={Diffusion-{LM} Improves Controllable Text Generation},
author={Xiang Lisa Li and John Thickstun and Ishaan Gulrajani and Percy Liang and Tatsunori Hashimoto},
booktitle={Advances in Neural Information Processing Systems},
editor={Alice H. Oh and Alekh Agarwal and Danielle Belgrave and Kyunghyun Cho},
year={2022},
url={https://openreview.net/forum?id=3s9IrEsjLyk}
}

@inproceedings{
ouyang2022instructgpt,
title={Training language models to follow instructions with human feedback},
author={Long Ouyang and Jeffrey Wu and Xu Jiang and Diogo Almeida and Carroll Wainwright and Pamela Mishkin and Chong Zhang and Sandhini Agarwal and Katarina Slama and Alex Gray and John Schulman and Jacob Hilton and Fraser Kelton and Luke Miller and Maddie Simens and Amanda Askell and Peter Welinder and Paul Christiano and Jan Leike and Ryan Lowe},
booktitle={Advances in Neural Information Processing Systems},
editor={Alice H. Oh and Alekh Agarwal and Danielle Belgrave and Kyunghyun Cho},
year={2022},
url={https://openreview.net/forum?id=TG8KACxEON}
}

@misc{schulman2017ppo,
      title={Proximal Policy Optimization Algorithms}, 
      author={John Schulman and Filip Wolski and Prafulla Dhariwal and Alec Radford and Oleg Klimov},
      year={2017},
      eprint={1707.06347},
      archivePrefix={arXiv},
      primaryClass={cs.LG},
      url={https://arxiv.org/abs/1707.06347}, 
}

@inproceedings{
rafailov2023dpo,
title={Direct Preference Optimization: Your Language Model is Secretly a Reward Model},
author={Rafael Rafailov and Archit Sharma and Eric Mitchell and Christopher D Manning and Stefano Ermon and Chelsea Finn},
booktitle={Thirty-seventh Conference on Neural Information Processing Systems},
year={2023},
url={https://openreview.net/forum?id=HPuSIXJaa9}
}

@article{zhao2025d1,
  title    = {d1: Scaling Reasoning in Diffusion Large Language Models via Reinforcement Learning},
  author   = {Zhao, Siyan and Gupta, Devaansh and Zheng, Qinqing and Grover, Aditya},
  journal  = {arXiv preprint arXiv:2504.12216},
  year     = {2025},
  doi      = {10.48550/arXiv.2504.12216}
}

@article{wang2025tracerl,
  title    = {Revolutionizing Reinforcement Learning Framework for Diffusion Large Language Models},
  author   = {Wang, Yinjie and Yang, Ling and Li, Bowen and Tian, Ye and Shen, Ke and Wang, Mengdi},
  journal  = {arXiv preprint arXiv:2509.06949},
  year     = {2025},
  doi      = {10.48550/arXiv.2509.06949}
}

@inproceedings{
wu2025fastdllm,
title={Fast-d{LLM}: Training-free Acceleration of Diffusion {LLM} by Enabling {KV} Cache and Parallel Decoding},
author={Chengyue Wu and Hao Zhang and Shuchen Xue and Zhijian Liu and Shizhe Diao and Ligeng Zhu and Ping Luo and Song Han and Enze Xie},
booktitle={The Fourteenth International Conference on Learning Representations},
year={2026},
url={https://openreview.net/forum?id=3Z3Is6hnOT}
}

@article{lu2025onpolicydistillation,
  author   = {Lu, Kevin and Thinking Machines Lab},
  title    = {On-Policy Distillation},
  journal  = {Thinking Machines Lab: Connectionism},
  year     = {2025},
  note     = {https://thinkingmachines.ai/blog/on-policy-distillation},
  doi      = {10.64434/tml.20251026}
}

@article{zhong2026stabledrl,
  title    = {Stabilizing Reinforcement Learning for Diffusion Language Models},
  author   = {Zhong, Jianyuan and Wang, Kaibo and Ding, Ding and Feng, Zijin and Bai, Haoli and Xiang, Yang and Sun, Jiacheng and Xu, Qiang},
  journal  = {arXiv preprint arXiv:2603.06743},
  year     = {2026},
  doi      = {10.48550/arXiv.2603.06743}
}

@inproceedings{
sahoo2024simple,
title={Simple and Effective Masked Diffusion Language Models},
author={Subham Sekhar Sahoo and Marianne Arriola and Aaron Gokaslan and Edgar Mariano Marroquin and Alexander M Rush and Yair Schiff and Justin T Chiu and Volodymyr Kuleshov},
booktitle={The Thirty-eighth Annual Conference on Neural Information Processing Systems},
year={2024},
url={https://openreview.net/forum?id=L4uaAR4ArM}
}

@misc{nie2025largelanguagediffusionmodels,
  title         = {Large Language Diffusion Models},
  author        = {Shen Nie and Fengqi Zhu and Zebin You and Xiaolu Zhang and Jingyang Ou and Jun Hu and Jun Zhou and Yankai Lin and Ji-Rong Wen and Chongxuan Li},
  year          = {2025},
  eprint        = {2502.09992},
  archiveprefix = {arXiv},
  primaryclass  = {cs.CL},
  url           = {https://arxiv.org/abs/2502.09992}
}

@article{guo2025deepseek,
   title={DeepSeek-R1 incentivizes reasoning in LLMs through reinforcement learning},
   volume={645},
   ISSN={1476-4687},
   url={http://dx.doi.org/10.1038/s41586-025-09422-z},
   DOI={10.1038/s41586-025-09422-z},
   number={8081},
   journal={Nature},
   publisher={Springer Science and Business Media LLC},
   author={Guo, Daya and Yang, Dejian and Zhang, Haowei and Song, Junxiao and Wang, Peiyi and Zhu, Qihao and Xu, Runxin and Zhang, Ruoyu and Ma, Shirong and Bi, Xiao and Zhang, Xiaokang and Yu, Xingkai and Wu, Yu and Wu, Z. F. and Gou, Zhibin and Shao, Zhihong and Li, Zhuoshu and Gao, Ziyi and Liu, Aixin and Xue, Bing and Wang, Bingxuan and Wu, Bochao and Feng, Bei and Lu, Chengda and Zhao, Chenggang and Deng, Chengqi and Ruan, Chong and Dai, Damai and Chen, Deli and Ji, Dongjie and Li, Erhang and Lin, Fangyun and Dai, Fucong and Luo, Fuli and Hao, Guangbo and Chen, Guanting and Li, Guowei and Zhang, H. and Xu, Hanwei and Ding, Honghui and Gao, Huazuo and Qu, Hui and Li, Hui and Guo, Jianzhong and Li, Jiashi and Chen, Jingchang and Yuan, Jingyang and Tu, Jinhao and Qiu, Junjie and Li, Junlong and Cai, J. L. and Ni, Jiaqi and Liang, Jian and Chen, Jin and Dong, Kai and Hu, Kai and You, Kaichao and Gao, Kaige and Guan, Kang and Huang, Kexin and Yu, Kuai and Wang, Lean and Zhang, Lecong and Zhao, Liang and Wang, Litong and Zhang, Liyue and Xu, Lei and Xia, Leyi and Zhang, Mingchuan and Zhang, Minghua and Tang, Minghui and Zhou, Mingxu and Li, Meng and Wang, Miaojun and Li, Mingming and Tian, Ning and Huang, Panpan and Zhang, Peng and Wang, Qiancheng and Chen, Qinyu and Du, Qiushi and Ge, Ruiqi and Zhang, Ruisong and Pan, Ruizhe and Wang, Runji and Chen, R. J. and Jin, R. L. and Chen, Ruyi and Lu, Shanghao and Zhou, Shangyan and Chen, Shanhuang and Ye, Shengfeng and Wang, Shiyu and Yu, Shuiping and Zhou, Shunfeng and Pan, Shuting and Li, S. S. and Zhou, Shuang and Wu, Shaoqing and Yun, Tao and Pei, Tian and Sun, Tianyu and Wang, T. and Zeng, Wangding and Liu, Wen and Liang, Wenfeng and Gao, Wenjun and Yu, Wenqin and Zhang, Wentao and Xiao, W. L. and An, Wei and Liu, Xiaodong and Wang, Xiaohan and Chen, Xiaokang and Nie, Xiaotao and Cheng, Xin and Liu, Xin and Xie, Xin and Liu, Xingchao and Yang, Xinyu and Li, Xinyuan and Su, Xuecheng and Lin, Xuheng and Li, X. Q. and Jin, Xiangyue and Shen, Xiaojin and Chen, Xiaosha and Sun, Xiaowen and Wang, Xiaoxiang and Song, Xinnan and Zhou, Xinyi and Wang, Xianzu and Shan, Xinxia and Li, Y. K. and Wang, Y. Q. and Wei, Y. X. and Zhang, Yang and Xu, Yanhong and Li, Yao and Zhao, Yao and Sun, Yaofeng and Wang, Yaohui and Yu, Yi and Zhang, Yichao and Shi, Yifan and Xiong, Yiliang and He, Ying and Piao, Yishi and Wang, Yisong and Tan, Yixuan and Ma, Yiyang and Liu, Yiyuan and Guo, Yongqiang and Ou, Yuan and Wang, Yuduan and Gong, Yue and Zou, Yuheng and He, Yujia and Xiong, Yunfan and Luo, Yuxiang and You, Yuxiang and Liu, Yuxuan and Zhou, Yuyang and Zhu, Y. X. and Huang, Yanping and Li, Yaohui and Zheng, Yi and Zhu, Yuchen and Ma, Yunxian and Tang, Ying and Zha, Yukun and Yan, Yuting and Ren, Z. Z. and Ren, Zehui and Sha, Zhangli and Fu, Zhe and Xu, Zhean and Xie, Zhenda and Zhang, Zhengyan and Hao, Zhewen and Ma, Zhicheng and Yan, Zhigang and Wu, Zhiyu and Gu, Zihui and Zhu, Zijia and Liu, Zijun and Li, Zilin and Xie, Ziwei and Song, Ziyang and Pan, Zizheng and Huang, Zhen and Xu, Zhipeng and Zhang, Zhongyu and Zhang, Zhen},
   year={2025},
   month=Sept, pages={633–638} }

@misc{shao2024deepseekmath,
      title={DeepSeekMath: Pushing the Limits of Mathematical Reasoning in Open Language Models}, 
      author={Zhihong Shao and Peiyi Wang and Qihao Zhu and Runxin Xu and Junxiao Song and Xiao Bi and Haowei Zhang and Mingchuan Zhang and Y. K. Li and Y. Wu and Daya Guo},
      year={2024},
      eprint={2402.03300},
      archivePrefix={arXiv},
      primaryClass={cs.CL},
      url={https://arxiv.org/abs/2402.03300}, 
}

@inproceedings{
agarwal2024policy,
title={On-Policy Distillation of Language Models: Learning from Self-Generated Mistakes},
author={Rishabh Agarwal and Nino Vieillard and Yongchao Zhou and Piotr Stanczyk and Sabela Ramos Garea and Matthieu Geist and Olivier Bachem},
booktitle={The Twelfth International Conference on Learning Representations},
year={2024},
url={https://openreview.net/forum?id=3zKtaqxLhW}
}

@misc{hinton2015distillingknowledgeneuralnetwork,
  title         = {Distilling the Knowledge in a Neural Network},
  author        = {Geoffrey Hinton and Oriol Vinyals and Jeff Dean},
  year          = {2015},
  eprint        = {1503.02531},
  archiveprefix = {arXiv},
  primaryclass  = {stat.ML},
  url           = {https://arxiv.org/abs/1503.02531}
}

@inproceedings{
chu2025sft,
title={{SFT} Memorizes, {RL} Generalizes: A Comparative Study of Foundation Model Post-training},
author={Tianzhe Chu and Yuexiang Zhai and Jihan Yang and Shengbang Tong and Saining Xie and Dale Schuurmans and Quoc V Le and Sergey Levine and Yi Ma},
booktitle={Forty-second International Conference on Machine Learning},
year={2025},
url={https://openreview.net/forum?id=dYur3yabMj}
}

@misc{inception2025mercury,
      title={Mercury: Ultra-Fast Language Models Based on Diffusion}, 
      author={Inception Labs and Samar Khanna and Siddhant Kharbanda and Shufan Li and Harshit Varma and Eric Wang and Sawyer Birnbaum and Ziyang Luo and Yanis Miraoui and Akash Palrecha and Stefano Ermon and Aditya Grover and Volodymyr Kuleshov},
      year={2025},
      eprint={2506.17298},
      archivePrefix={arXiv},
      primaryClass={cs.CL},
      url={https://arxiv.org/abs/2506.17298}, 
}

@inproceedings{arriola2025block,
  title     = {Block Diffusion: Interpolating Between Autoregressive and Diffusion Language Models},
  author    = {Arriola, Marianne and Gokaslan, Aaron and Chiu, Justin T and Yang, Zhihan and Qi, Zhixuan and Han, Jiaqi and Sahoo, Subham Sekhar and Kuleshov, Volodymyr},
  booktitle = {The Thirteenth International Conference on Learning Representations},
  year      = {2025},
  url       = {https://arxiv.org/abs/2503.09573}
}

@inproceedings{
shi2024simplified,
title={Simplified and Generalized Masked Diffusion for Discrete Data},
author={Jiaxin Shi and Kehang Han and Zhe Wang and Arnaud Doucet and Michalis Titsias},
booktitle={The Thirty-eighth Annual Conference on Neural Information Processing Systems},
year={2024},
url={https://openreview.net/forum?id=xcqSOfHt4g}
}

@inproceedings{
ou2025espo,
title={Principled {RL} for Diffusion {LLM}s Emerges from a Sequence-Level Perspective},
author={Jingyang Ou and Jiaqi Han and Minkai Xu and Shaoxuan Xu and Jianwen Xie and Stefano Ermon and Yi Wu and Chongxuan Li},
booktitle={The Fourteenth International Conference on Learning Representations},
year={2026},
url={https://openreview.net/forum?id=S5YeC9llIL}
}

@inproceedings{
wang2025spg,
title={{SPG}: Sandwiched Policy Gradient for Masked Diffusion Language Models},
author={Chenyu Wang and Paria Rashidinejad and DiJia Su and Song Jiang and Sid Wang and Siyan Zhao and Cai Zhou and Shannon Zejiang Shen and Feiyu Chen and Tommi Jaakkola and Yuandong Tian and Bo Liu},
booktitle={The Fourteenth International Conference on Learning Representations},
year={2026},
url={https://openreview.net/forum?id=18j5Q49GwN}
}

@article{zhao2025diffpo,
  title         = {DiFFPO: Training Diffusion LLMs to Reason Fast and Furious via Reinforcement Learning},
  author        = {Zhao, Hanyang and Liang, Dawen and Tang, Wenpin and Yao, David D. and Kallus, Nathan},
  journal       = {arXiv preprint arXiv:2510.02212},
  year          = {2025},
  eprint        = {2510.02212},
  archiveprefix = {arXiv},
  primaryclass  = {cs.LG}
}

@article{schulman2015trpo,
  title         = {Trust Region Policy Optimization},
  author        = {Schulman, John and Levine, Sergey and Moritz, Philipp and Jordan, Michael I. and Abbeel, Pieter},
  journal       = {arXiv preprint arXiv:1502.05477},
  year          = {2015},
  eprint        = {1502.05477},
  archiveprefix = {arXiv},
  primaryclass  = {cs.LG}
}

@inproceedings{
gong2025scaling,
title={Scaling Diffusion Language Models via Adaptation from Autoregressive Models},
author={Shansan Gong and Shivam Agarwal and Yizhe Zhang and Jiacheng Ye and Lin Zheng and Mukai Li and Chenxin An and Peilin Zhao and Wei Bi and Jiawei Han and Hao Peng and Lingpeng Kong},
booktitle={The Thirteenth International Conference on Learning Representations},
year={2025},
url={https://openreview.net/forum?id=j1tSLYKwg8}
}

@misc{cheng2025sdar,
      title={SDAR: A Synergistic Diffusion-AutoRegression Paradigm for Scalable Sequence Generation}, 
      author={Shuang Cheng and Yihan Bian and Dawei Liu and Linfeng Zhang and Qian Yao and Zhongbo Tian and Wenhai Wang and Qipeng Guo and Kai Chen and Biqing Qi and Bowen Zhou},
      year={2025},
      eprint={2510.06303},
      archivePrefix={arXiv},
      primaryClass={cs.LG},
      url={https://arxiv.org/abs/2510.06303}, 
}

@misc{ross2011reduction,
      title={A Reduction of Imitation Learning and Structured Prediction to No-Regret Online Learning}, 
      author={Stephane Ross and Geoffrey J. Gordon and J. Andrew Bagnell},
      year={2011},
      eprint={1011.0686},
      archivePrefix={arXiv},
      primaryClass={cs.LG},
      url={https://arxiv.org/abs/1011.0686}, 
}

@misc{cobbe2021training,
      title={Training Verifiers to Solve Math Word Problems}, 
      author={Karl Cobbe and Vineet Kosaraju and Mohammad Bavarian and Mark Chen and Heewoo Jun and Lukasz Kaiser and Matthias Plappert and Jerry Tworek and Jacob Hilton and Reiichiro Nakano and Christopher Hesse and John Schulman},
      year={2021},
      eprint={2110.14168},
      archivePrefix={arXiv},
      primaryClass={cs.LG},
      url={https://arxiv.org/abs/2110.14168}, 
}

@misc{sanh2019distilbert,
      title={DistilBERT, a distilled version of BERT: smaller, faster, cheaper and lighter}, 
      author={Victor Sanh and Lysandre Debut and Julien Chaumond and Thomas Wolf},
      year={2020},
      eprint={1910.01108},
      archivePrefix={arXiv},
      primaryClass={cs.CL},
      url={https://arxiv.org/abs/1910.01108}, 
}

@misc{song2026survey,
      title={A Survey on Knowledge Distillation of Large Language Models}, 
      author={Xiaohan Xu and Ming Li and Chongyang Tao and Tao Shen and Reynold Cheng and Jinyang Li and Can Xu and Dacheng Tao and Tianyi Zhou},
      year={2024},
      eprint={2402.13116},
      archivePrefix={arXiv},
      primaryClass={cs.CL},
      url={https://arxiv.org/abs/2402.13116}, 
}

@inproceedings{
xuspeculative,
title={Speculative Knowledge Distillation: Bridging the Teacher-Student Gap Through Interleaved Sampling},
author={Wenda Xu and Rujun Han and Zifeng Wang and Long Le and Dhruv Madeka and Lei Li and William Yang Wang and Rishabh Agarwal and Chen-Yu Lee and Tomas Pfister},
booktitle={The Thirteenth International Conference on Learning Representations},
year={2025},
url={https://openreview.net/forum?id=EgJhwYR2tB}
}

@inproceedings{
gu2024minillm,
title={Mini{LLM}: Knowledge Distillation of Large Language Models},
author={Yuxian Gu and Li Dong and Furu Wei and Minlie Huang},
booktitle={The Twelfth International Conference on Learning Representations},
year={2024},
url={https://openreview.net/forum?id=5h0qf7IBZZ}
}

@inproceedings{
hendrycks2021measuring,
title={Measuring Mathematical Problem Solving With the {MATH} Dataset},
author={Dan Hendrycks and Collin Burns and Saurav Kadavath and Akul Arora and Steven Basart and Eric Tang and Dawn Song and Jacob Steinhardt},
booktitle={Thirty-fifth Conference on Neural Information Processing Systems Datasets and Benchmarks Track (Round 2)},
year={2021},
url={https://openreview.net/forum?id=7Bywt2mQsCe}
}

@inproceedings{
hu2025open,
title={Open-Reasoner-Zero: An Open Source Approach to Scaling Up Reinforcement Learning on the Base Model},
author={Jingcheng Hu and Yinmin Zhang and Qi Han and Daxin Jiang and Xiangyu Zhang and Heung-Yeung Shum},
booktitle={The Thirty-ninth Annual Conference on Neural Information Processing Systems},
year={2026},
url={https://openreview.net/forum?id=NFM8F5cV0V}
}

@misc{maa2024aime,
  title        = {American Invitational Mathematics Examination (AIME) 2024: AIME I and AIME II},
  author       = {{Mathematical Association of America, American Mathematics Competitions}},
  year         = {2024},
  howpublished = {\url{https://artofproblemsolving.com/wiki/index.php/AIME_Problems_and_Solutions}},
  note         = {Competition problems used as an evaluation dataset; original problems by MAA AMC},
  urldate      = {2025-09-01},
  organization = {Mathematical Association of America (AMC)}
}

@misc{yu2025dapo,
      title={DAPO: An Open-Source LLM Reinforcement Learning System at Scale}, 
      author={Qiying Yu and Zheng Zhang and Ruofei Zhu and Yufeng Yuan and Xiaochen Zuo and Yu Yue and Weinan Dai and Tiantian Fan and Gaohong Liu and Lingjun Liu and Xin Liu and Haibin Lin and Zhiqi Lin and Bole Ma and Guangming Sheng and Yuxuan Tong and Chi Zhang and Mofan Zhang and Wang Zhang and Hang Zhu and Jinhua Zhu and Jiaze Chen and Jiangjie Chen and Chengyi Wang and Hongli Yu and Yuxuan Song and Xiangpeng Wei and Hao Zhou and Jingjing Liu and Wei-Ying Ma and Ya-Qin Zhang and Lin Yan and Mu Qiao and Yonghui Wu and Mingxuan Wang},
      year={2025},
      eprint={2503.14476},
      archivePrefix={arXiv},
      primaryClass={cs.LG},
      url={https://arxiv.org/abs/2503.14476}, 
}

@article{zhu2025llada,
  title         = {{LLaDA} 1.5: Variance-Reduced Preference Optimization for Large Language Diffusion Models},
  author        = {Zhu, Fengqi and Wang, Rongzhen and Nie, Shen and Zhang, Xiaolu and Wu, Chunwei and Hu, Jun and Zhou, Jun and Chen, Jianfei and Lin, Yankai and Wen, Ji-Rong and Li, Chongxuan},
  journal       = {arXiv preprint arXiv:2505.19223},
  year          = {2025},
  eprint        = {2505.19223},
  archiveprefix = {arXiv},
  primaryclass  = {cs.CL}
}

@article{he2025mdpo,
  title         = {{MDPO}: Overcoming the Training-Inference Divide of Masked Diffusion Language Models},
  author        = {He, Haoyu and Renz, Katrin and Cao, Yong and Geiger, Andreas},
  journal       = {arXiv preprint arXiv:2508.13148},
  year          = {2025},
  eprint        = {2508.13148},
  archiveprefix = {arXiv},
  primaryclass  = {cs.LG},
  doi           = {10.48550/arXiv.2508.13148}
}

@misc{bie2025llada20scalingdiffusionlanguage,
  title         = {LLaDA2.0: Scaling Up Diffusion Language Models to 100B},
  author        = {Tiwei Bie and Maosong Cao and Kun Chen and Lun Du and Mingliang Gong and Zhuochen Gong and Yanmei Gu and Jiaqi Hu and Zenan Huang and Zhenzhong Lan and Chengxi Li and Chongxuan Li and Jianguo Li and Zehuan Li and Huabin Liu and Lin Liu and Guoshan Lu and Xiaocheng Lu and Yuxin Ma and Jianfeng Tan and Lanning Wei and Ji-Rong Wen and Yipeng Xing and Xiaolu Zhang and Junbo Zhao and Da Zheng and Jun Zhou and Junlin Zhou and Zhanchao Zhou and Liwang Zhu and Yihong Zhuang},
  year          = {2025},
  eprint        = {2512.15745},
  archiveprefix = {arXiv},
  primaryclass  = {cs.LG},
  url           = {https://arxiv.org/abs/2512.15745}
}

@misc{ye2025dream7bdiffusionlarge,
  title         = {Dream 7B: Diffusion Large Language Models},
  author        = {Jiacheng Ye and Zhihui Xie and Lin Zheng and Jiahui Gao and Zirui Wu and Xin Jiang and Zhenguo Li and Lingpeng Kong},
  year          = {2025},
  eprint        = {2508.15487},
  archiveprefix = {arXiv},
  primaryclass  = {cs.CL},
  url           = {https://arxiv.org/abs/2508.15487}
}

@misc{song2026surveyonpolicydistillationlarge,
  title         = {A Survey of On-Policy Distillation for Large Language Models},
  author        = {Mingyang Song and Mao Zheng},
  year          = {2026},
  eprint        = {2604.00626},
  archiveprefix = {arXiv},
  primaryclass  = {cs.LG},
  url           = {https://arxiv.org/abs/2604.00626}
}
